\documentclass{article} 
\usepackage{iclr2026_conference,times}

\usepackage{amsmath,amsfonts,bm}

\def\eqref#1{equation~\ref{#1}}

\def\1{\bm{1}}

\DeclareMathAlphabet{\mathsfit}{\encodingdefault}{\sfdefault}{m}{sl}
\SetMathAlphabet{\mathsfit}{bold}{\encodingdefault}{\sfdefault}{bx}{n}

\usepackage{hyperref}
\usepackage{url}
\usepackage{graphicx}
\usepackage{subcaption}
\usepackage{array}
\usepackage{algorithm}
\usepackage{tabularx}
\usepackage{booktabs}   
\usepackage{multirow}
\usepackage{wrapfig}
\usepackage{needspace}
\usepackage[noend]{algpseudocode}
\usepackage[table]{xcolor}
\usepackage{wasysym}   
\usepackage{amssymb}   
\usepackage{wrapfig}
\usepackage{enumitem}

\usepackage{xcolor}
\usepackage[framemethod=tikz]{mdframed}

\usepackage[most]{tcolorbox}
\tcbuselibrary{listings,skins,breakable} 

\title{Deep Literature Survey Automation with an Iterative Workflow}

\newcommand{\au}[2]{\textbf{#1}\textsuperscript{#2}} 

\author{%
\au{Hongbo Zhang}{1,2}$^*$,\quad \au{Han Cui}{1,2}\thanks{Equal contribution.},\quad \au{Yidong Wang}{2,3},\quad \au{Yijian Tian}{2},\quad \\
\au{Qi Guo}{3},\quad \au{Cunxiang Wang}{2},\quad  \au{Jian Wu}{2},\quad \au{Chiyu Song}{2},\quad \au{Yue Zhang}{2,4}\thanks{Corresponding author.}
\\[3pt]
\normalsize
\textsuperscript{1}\,Zhejiang University\quad
\textsuperscript{2}\,School of Engineering, Westlake University \quad
\textsuperscript{3}\,Peking University \quad\\
\textsuperscript{4}\,Institute of Advanced Technology, Westlake Institute for Advanced Study 
\\[2pt]
\texttt{\{zhanghongbo,cuihan,zhangyue\}@westlake.edu.cn} \quad
}

\newcommand{\as}{AutoSurvey}
\newcommand{\ours}{IterSurvey}

\iclrfinalcopy

\begin{document}

\maketitle

\begin{abstract}
Automatic literature survey generation has attracted increasing attention, yet most existing systems follow a one-shot paradigm, where a large set of papers is retrieved at once and a static outline is generated before drafting. This design often leads to noisy retrieval, fragmented structures, and context overload, ultimately limiting survey quality. Inspired by the iterative reading process of human researchers, we propose \ours, a framework based on recurrent outline generation, in which a planning agent incrementally retrieves, reads, and updates the outline to ensure both exploration and coherence. To provide faithful paper-level grounding, we design paper cards that distill each paper into its contributions, methods, and findings, and introduce a review-and-refine loop with visualization enhancement to improve textual flow and integrate multimodal elements such as figures and tables. Experiments on both established and emerging topics show that \ours\ substantially outperforms state-of-the-art baselines in content coverage, structural coherence, and citation quality, while producing more accessible and better-organized surveys. To provide a more reliable assessment of such improvements, we further introduce Survey-Arena, a pairwise benchmark that complements absolute scoring and more clearly positions machine-generated surveys relative to human-written ones. The code is available at \url{https://github.com/HancCui/IterSurvey\_Autosurveyv2}.

\end{abstract}

\section{Introduction}

Automatic literature survey generation has recently attracted growing attention due to its potential to help researchers quickly grasp new domains, identify key trends, and reduce the burden of manual reviews. Following ~\citet{wang2024autosurvey}, current systems generally adopt a multistage pipeline~\citep{liang2025surveyx, yan-etal-2025-surveyforge,wang2025llm}: The process begins with a topic description, usually consisting of a few tokens, which is directly used to retrieve a large collection of candidate papers. 
Due to the context window limitation of large language models (LLMs), the retrieved papers are divided into multiple groups, for each, an LLM agent generates a survey section outline based on the corresponding subset of papers. 
These group-level outlines are subsequently merged into a global draft outline. 
Once the draft outline is obtained, the system performs section-wise retrieval to collect references for section writing and then generates the corresponding text passages. 
Finally, a global review and integration process is applied, in which the drafted survey is iteratively polished to improve readability and overall consistency.

The above approach takes a "one-shot" planning paradigm, retrieves a comprehensive set of papers and construct a global outline from a single, static starting point. This approach, however, leads to several limitations. 
\textbf{First, retrieval can be imprecise and static} due to reliance on a short topic description (often just a few tokens) as the retrieval query~\citep{sun-etal-2019-pullnet,AZAD20191698,wang2020deepreinforcedqueryreformulation}. Such coarse queries fail to capture a field's nuances and are never refined, leading to noisy and incomplete paper collections. 
\textbf{Second, the survey structure can be incoherent}~\citep{fabbri-etal-2019-multi,gidiotis2020divideandconquerapproachsummarizationlong, yang-etal-2023-doc}. 
Since outlines are generated for each paper group independently and subsequently merged, the global structure lacks coherence and often misses important cross-group connections. 
\textbf{Third, injecting overly long contexts introduces distraction and context overload}~\citep{liu2023lostmiddlelanguagemodels, wu2024reducingdistractionlongcontextlanguage}.
Feeding entire papers into LLMs not only exposes them to large amounts of peripheral information, such as dataset details or experimental setups, which distracts from the conceptual structure needed for survey writing, but also places unnecessary pressure on the limited context window of the model.

\begin{figure}
    \centering
    \includegraphics[width=1.0\linewidth]{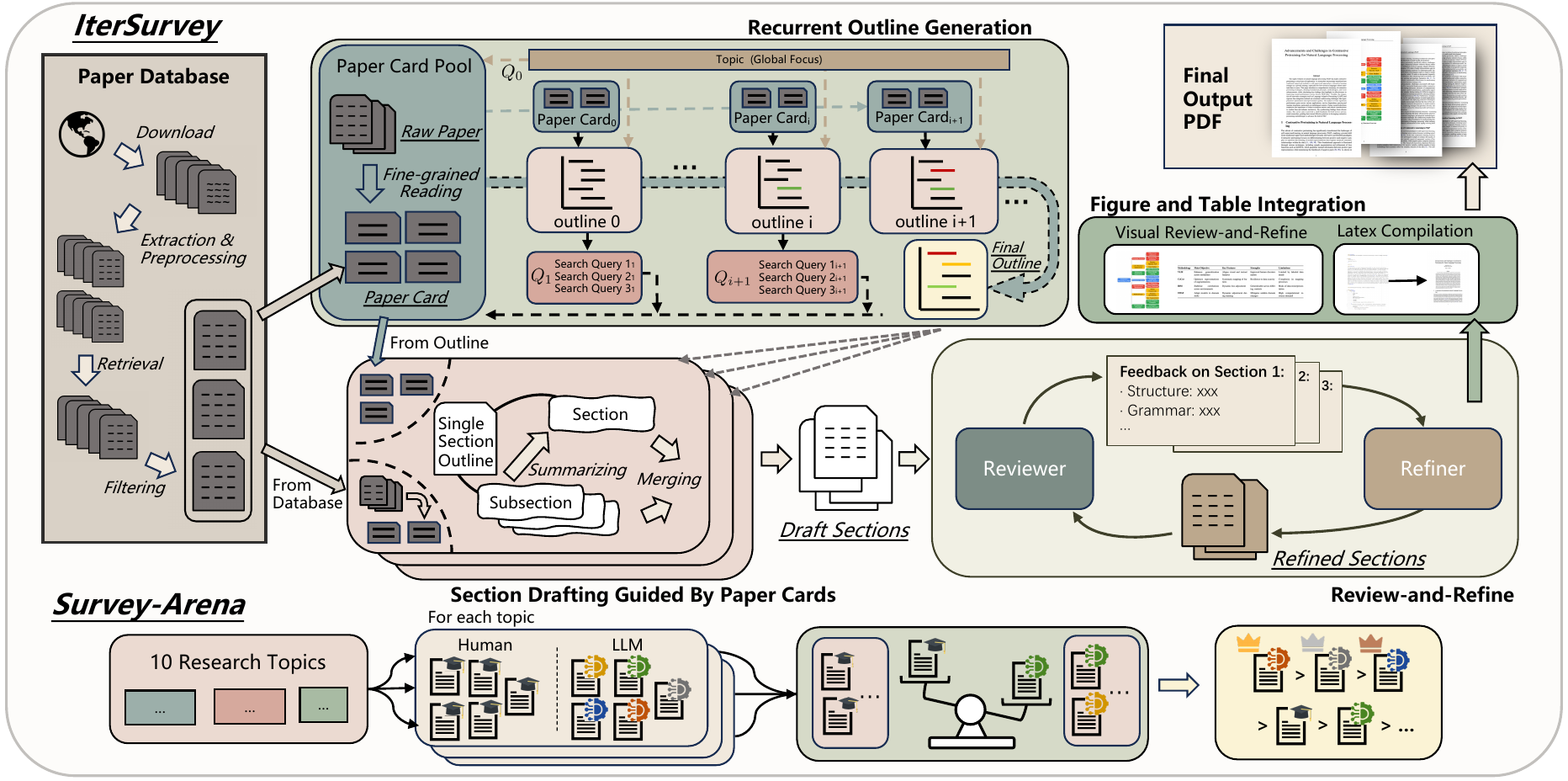}
    \caption{Overview of IterSurvey and Survey-Arena.}
    \label{fig:overview}
    \vspace{-1.5em}
\end{figure}
In contrast, human researchers rarely attempt to grasp an entire field in a single shot. Instead, they follow an iterative reading process: starting with a small set of core papers, summarizing key contributions, and gradually expanding to related directions as their understanding deepens~\citep{bates1989berrypicking, asai2023selfraglearningretrievegenerate}.
Inspired by this workflow, we propose an iterative planning paradigm for automated survey generation. At its core lies a \textbf{recurrent outline generation} module that incrementally retrieves, organizes, and integrates evidence through a planning agent equipped with stability checks and stopping criteria, mitigating the brittleness of one-shot pipelines that rely on static queries and fragmented merges. Central to this process are \textbf{paper cards}, structured semantic abstractions that distill each paper into contributions, methods, and findings. Unlike conventional abstract-based inputs, these cards serve as fine-grained evidence units that guide both outline construction and \textbf{section drafting}, ensuring coherence and faithful citation across iterations. Finally, a \textbf{global review and integration} stage employs a reviewer–refiner loop to enforce consistency and clarity across sections, while an integrated figure–table generation pipeline compiles candidate visualizations, automatically checks them for layout and readability, and revises them to meet academic presentation standards. 
This design inherits the advantages of iterative human reading: retrieval is progressively refined rather than static~\citep{jiang2023activeretrievalaugmentedgeneration}, the outline develops as an organically coherent structure rather than a patchwork~\citep{zhang2025evolvesearchiterativeselfevolvingsearch}, and paper cards enforce fine-grained evidence grounding that avoids distraction from peripheral details~\citep{cachola2020tldrextremesummarizationscientific, wu2024reducingdistractionlongcontextlanguage}.

Comprehensive experiments validate the effectiveness of our incremental paradigm. \ours\ consistently outperforms all baselines across multiple dimensions, with recurrent outline generation yielding more coherent structures and paper cards improving citation accuracy without sacrificing precision. These advantages are further confirmed by human evaluation, where experts also favor the outputs of \ours\ over competing systems. While these results confirm the superiority of \ours, we find that absolute scoring struggles to reliably quantify the performance gap against human-written surveys~\citep{yang2023rethinkingbenchmarkcontaminationlanguage,oren2023provingtestsetcontamination, ye2024yetrevealingrisksutilizing}.  In the LLM evaluation community, similar concerns have led to the development of Chatbot Arena~\cite{chiang2024chatbotarenaopenplatform}, which adopts pairwise human preference judgments to overcome the noisiness and inconsistency of absolute ratings. Inspired by this paradigm, we further contribute \textbf{Survey-Arena}, the first benchmark to our knowledge that evaluates synthesized surveys through direct, pairwise ranking against a corpus of human-written exemplars. This approach provides a more robust and interpretable assessment of system quality by directly positioning it relative to a human-level baseline.

Our contributions are threefold.
\vspace{-0.5em}
\begin{itemize}
    \item We propose \textbf{recurrent outline generation}, which iteratively retrieves, reads, and updates outlines with paper cards and outline–paper grounding, while encouraging the model to explore new directions. 
    \item We develop a new framework: \textbf{\ours}, which produces finer-grained outlines and supports multi-modal inputs and outputs for more comprehensive surveys.  
    \item We construct \textbf{Survey-Arena}, a pairwise evaluation benchmark that ranks machine-generated surveys alongside human-written ones, enabling more reliable and interpretable assessment of survey quality.
\end{itemize}

\section{Related Work}

\paragraph{Automated Survey Generation}
Recent automated survey generation systems largely adopt a "one-shot" paradigm, where a static outline is constructed upfront before content generation. This approach is evident in pipeline-based systems like AutoSurvey~\citep{wang2024autosurvey}, which employs a hierarchical paradigm, and SurveyForge~\citep{yan-etal-2025-surveyforge}, which utilizes a memory-driven scholar navigation agent. Other frameworks focus on enhancing this initial outlining step through reference pre-processing; for instance, SurveyX~\citep{liang2025surveyx} introduces an AttributeTree to extract key information, while HiReview~\citep{hu2024hireview} generates a hierarchical taxonomy tree. Tackling the challenge from a technical scalability perspective, SurveyGo~\citep{wang2025llm} leverages the LLM×MapReduce-V2 algorithm to handle long contexts within this paradigm. In contrast, our framework treats the outline not as a static blueprint but as an evolving knowledge structure. Through a dynamic, recurrent mechanism, the outline is continuously updated as the system iteratively engages with the literature, resulting in comprehensive and coherent synthesis.

\paragraph{Evaluation of Automated Surveys}
Evaluating machine-generated surveys is inherently challenging. Building on insights from automated peer review~\citep{yu2024automatedpeerreviewingpaper, jin2024agentreviewexploringpeerreview, weng2025cycleresearcherimprovingautomatedresearch}, prior works~\citep{wang2024autosurvey, yan-etal-2025-surveyforge, liang2025surveyx} commonly adopt an LLM-as-a-judge paradigm with manually designed criteria, assessing dimensions such as coherence, coverage, and factuality. Citation quality is typically measured with NLI-based protocols~\citep{gao2023enabling}, and \citet{yan-etal-2025-surveyforge} additionally evaluate coverage by comparing system outputs with human-written surveys. While absolute scoring by LLMs provides useful fine-grained signals, it has also been noted to suffer from inconsistency and calibration issues~\citep{ye2024yetrevealingrisksutilizing, latona2024aireviewlotterywidespread}, making system-level comparisons less reliable. In contrast, pairwise judgment which is widely used in chatbot evaluation~\citep{SE_arena, chiang2024chatbotarenaopenplatform} and peer review~\citep{zhang2025replicationredesignexploringpairwise}, offers more stable and interpretable assessments, but has not yet been applied to survey evaluation. To fill this gap, we introduce \textit{Survey-Arena}, the first benchmark that ranks machine-generated surveys against human-written exemplars, providing both robust comparison across systems and a clearer positioning relative to human-level quality.

\section{IterSurvey}
An overview of \ours\ is shown in Fig.~\ref{fig:overview}, and its three core stages are detailed below.

\subsection{Recurrent Outline Generation}
\label{subsec:recurrent_outline_generation}
\begin{algorithm}[t]\scriptsize
\begin{algorithmic}[1]
\Require Topic query $q$; retrieval sizes $(n,m)$; batch size $B$; paper budget $(N_{\min},N_{\max})$; similarity threshold $\tau$
\Ensure Writing-oriented outline $\hat{O}$

\State $O \gets \textsc{InitOutline}(q)$
\State $\mathsf{Pool} \gets \emptyset$ \Comment{map: query $\mapsto$ card list}
\State $\mathcal{U} \gets \emptyset$ \Comment{consulted papers}
\State $\mathbf{R} \gets [\,]$ \Comment{query history}

\ForAll{$r \in \textsc{SeedQueries}(q)$}
  \State $\mathcal{P} \gets \textsc{Retrieve}(r,n) \cup \textsc{TopRefs}(\cdot,m)$
  \State $\mathcal{C} \gets \{ \textsc{PaperCard}(p)\mid p\in\mathcal{P} \}$
  \State $\mathsf{Pool}[r] \gets \mathcal{C}$; \quad $\mathcal{U} \gets \mathcal{U} \cup \mathcal{P}$
\EndFor

\While{$|\mathcal{U}| < N_{\max}$}
  \If{$\mathsf{Pool}=\emptyset$}
    \If{$|\mathcal{U}| \ge N_{\min}$ \textbf{ and } $h(O,\mathbf{R})$}
      \State \textbf{break}
    \Else
      \ForAll{$r \in \textsc{ExpandQueries}(O,\mathbf{R})$}
        \State $\mathcal{P} \gets \textsc{Retrieve}(r,n) \cup \textsc{TopRefs}(\cdot,m)$
        \State $\mathcal{C} \gets \{ \textsc{PaperCard}(p)\mid p\in\mathcal{P} \}$
        \State $\mathsf{Pool}[r] \gets \mathcal{C}$; \quad $\mathcal{U} \gets \mathcal{U} \cup \mathcal{P}$
      \EndFor
      \State \textbf{continue}
    \EndIf
  \EndIf

  \State $(r,\mathcal{C}) \gets \textsc{Pop}(\mathsf{Pool})$ \Comment{activate a query and its cards}
  \State $\mathbf{R} \gets \mathbf{R} \,\|\, r$
  \While{$\mathcal{C} \neq \emptyset$}
    \State $\mathcal{B} \gets \textsc{SampleBatch}(\mathcal{C}, B)$
    \State $\tilde{O} \gets g(O,\mathcal{B}, r)$ \Comment{retrieval + reading + synthesis}
    \If{$\textsc{Sim}(O,\tilde{O}) \ge \tau$} \State $O \gets \tilde{O}$ \EndIf
    \State $\mathcal{C} \gets \mathcal{C} \setminus \mathcal{B}$
  \EndWhile
\EndWhile

\State $\hat{O} \gets \textsc{Refine}(O)$
\State \Return $\hat{O}$
\end{algorithmic}
\caption{Description of the recurrent outline generation process.}
\label{alg:rnn_outline_pseudocode}
\end{algorithm}
Outline generation is a central component of automatic survey construction, as it requires understanding the research domain, identifying its subfields, and synthesizing individual papers. Alg.~\ref{alg:rnn_outline_pseudocode} shows the overview of the generation process. The outcome is a hierarchical framework that summarizes the domain, where each node in the hierarchy is represented by a title and an accompanying description. 
Given a topic query, our goal is to enable the model to integrate retrieval with inductive reasoning, so that it can systematically explore the literature and produce a comprehensive outline for the target domain. 
To this end, we design recurrent outline generation.

\paragraph{Paper Card Pool.} 
The paper card pool organizes retrieval keywords together with their associated papers in a structured mapping.
For each keyword $K_i$, we retrieve $n$ candidate papers and extract $m$ of the most relevant references, forming the set:
\[
\mathcal{P}_i = \{p_i^1, p_i^2, \dots, p_i^{n+m}\}.
\]
At iteration $i$, the system pops one keyword $K_i$ together with its associated paper set $\mathcal{P}_i$ from the pool. 
Each paper $p_i^j \in \mathcal{P}_i$ is converted into a paper card
\[
c_i^j = \texttt{PaperCard}(p_i^j),
\]
which distills the paper into its contributions, methods, and findings. 
The collection of paper cards is denoted as $\mathcal{C}_i = \{c_i^1, c_i^2, \dots, c_i^{|\mathcal{P}_i|}\}$. 
Overall, the paper card pool can be represented as a mapping
\[
\mathcal{Q} = \{\, K_i \mapsto \mathcal{C}_i \mid i=0,1,\dots \,\},
\]
where each keyword $K_i$ is associated with the corresponding set of paper cards $\mathcal{C}_i$.

\paragraph{Outline updating.}
The outline updating process begins with an empty initial outline, denoted as $O_0$. At each step, the outline is refined using the current outline $O_i$, the active keyword $K_i$, and a mini-batch of paper cards drawn from the pool. 
Specifically, let $\mathcal{B}_i \subseteq \mathcal{C}_i$ be a batch of paper cards sampled from the set of cards associated with $K_i$. 
The model produces a candidate update
\[
\tilde{O}_{i+1} = g(O_i, \mathcal{B}_i, K_i),
\]
where $g(\cdot)$ denotes the outline updating function.
This procedure is repeated iteratively, with batches $\mathcal{B}_i$ of paper cards popped from the paper pool $\mathcal{Q}$ under the current keyword $K_i$, until all cards associated with $K_i$ are consumed and integrated into the outline.
To ensure stability and promote refinement, the candidate update is accepted if its similarity to the previous outline exceeds $\tau$:
\[
O_{i+1} =
\begin{cases}
\tilde{O}_{i+1}, & \text{if } \texttt{Sim}(O_i, \tilde{O}_{i+1}) \geq \tau, \\
O_i, & \text{otherwise}.
\end{cases}
\]

\paragraph{Keyword expansion.}
When all keywords $K_i$ has been fully consumed, the system explores new directions by proposing additional keywords. The goal is to identify potentially relevant aspects of the domain that have not yet been covered. 
Formally, new keywords are generated as
\[
K_{i+1} = f(O_{i+1}, K_i, \dots, K_0),
\]
where $f(\cdot)$ denotes a keyword generation function that takes the updated outline and the history of queries as input, and proposes candidate keywords for further exploration.
The corresponding paper set $\mathcal{P}_{i+1}$ is then retrieved and pushed into the pool $\mathcal{Q}$, thereby guiding the next iteration.

\paragraph{Stopping condition.}
Let $N_i = |\mathcal{P}_0 \cup \mathcal{P}_1 \cup \dots \cup \mathcal{P}_i|$ denote the total number of consulted papers up to iteration $i$. 
The process terminates when either (i) $N_i \geq N_{\min}$ and the stopping signal
\[
s = h(O_{i+1}, K_i, \dots, K_0), \quad s \in \{0,1\},
\]
indicates that the outline is sufficiently complete, or (ii) $N_i \geq N_{\max}$. 
Here $h(\cdot)$ is a decision function which takes the evolving outline and the query history as input and outputs whether further exploration is necessary. 
This design ensures that the outline is not terminated prematurely, while also preventing excessive exploration.

\paragraph{Post-processing.}
After termination, the recurrent process produces a research-oriented outline $\tilde{O}$, which is further refined into a writing-oriented survey outline:
\[
\hat{O} = \texttt{Refine}(\tilde{O}),
\]
where $\texttt{Refine}(\cdot)$ reorganizes the structure, inserts standard survey components such as `Introduction' and `Future Directions', and ensures conformity with academic conventions. Finally, we perform paper–section relinking, where all consulted papers are reassociated with the corresponding sections of the final outline $\hat{O}$. 
This guarantees that each section of $\hat{O}$ is grounded in concrete evidence, providing a reliable foundation for subsection drafting.

\subsection{Section Drafting Guided by Paper Cards}
\label{subsec:section_drafting}
A distinctive feature of our framework is that section drafting is entirely guided by paper cards, which serve as fine-grained, structured representations of the literature. 
Given the refined outline $\hat{O}$, each section or subsection is written by conditioning on its description $d_j$ together with the relevant pool of cards. 
Specifically, for a given subsection with description $d_j$, the system retrieves a set of additional reference papers $\mathcal{P}_{\text{sec}}^j$ and converts them into paper cards $\mathcal{C}_{\text{sec}}^j$. 
In contrast to previous work, our framework benefits from the paper–section relinking established during outline construction: each subsection is already associated with a pool of consulted papers from earlier iterations. 
This enriched evidence base, combining $\mathcal{C}_{\text{sec}}^j$ with the relinked cards, provides the model with a stronger foundation for subsection writing. Formally, the $j$-th subsection is generated as
\[
S_j = \texttt{Draft}(d_j, \mathcal{C}_{\text{sec}}^j \cup \mathcal{C}_{\text{link}}^j),
\]
where $\mathcal{C}_{\text{link}}^j$ denotes the set of paper cards relinked to subsection $j$. 
During drafting, the model is required to cite the provided references, and the citations are mapped to their corresponding papers.

\subsection{Global Review and Integration}
\label{subsec:review_refine}
The final stage of survey generation goes beyond local drafting. It performs a global review-and-refine process that integrates sections into a coherent survey and enriches the survey with automatically generated figures and tables.

\paragraph{Textual Review-and-Refine.}
We adopt a reviewer–refiner loop that involves two collaborative LLM roles. The reviewer takes the entire survey draft as input to capture the global context but then focuses its critique on a specific section or subsection. This design ensures that feedback on local content is always grounded in an understanding of the overall narrative. The reviewer provides detailed suggestions covering aspects such as clarity of exposition, consistency of terminology, logical alignment with preceding and following sections, and stylistic fluency. The refiner then incorporates these suggestions to revise the targeted section, producing a polished update that fits better into the survey as a whole. This loop is applied sequentially across all sections and iterated multiple times, progressively enhancing readability, improving cross-section coherence, and strengthening the global structural integrity of the survey.

\paragraph{Figure–Table Integration.}
In addition to textual refinement, we extend the refinement process to include multimodal elements, to further enhance readability.
For each section, the model first generates visualization requirements, such as tables with structured comparisons or figures with explanatory diagrams, together with natural language descriptions.
Based on these descriptions, candidate figures and tables are synthesized. 
The compiled outputs are then fed back to an LLM for quality assessment, enabling automatic detection of issues such as oversized layouts or unreadable text. 
The LLM provides corrective suggestions, which are applied to improve the final visualizations. 
Finally, the text is refined again to ensure that all generated figures and tables are properly referenced within the survey.

\section{Experiments}

\subsection{Experimental Settings}
\paragraph{Implementation Details.}  

Following\ ~\citet{wang2024autosurvey}, we adopt \textbf{GPT-4o-mini} as our generation model for its balance of responsiveness and cost.
Our retrieval database contains 680K computer science papers from arXiv, with PDFs converted into structured Markdown using MinerU~\citep{wang2024mineru} for consistent formatting. The details of the retrieval process are provided in App.~\ref{app:retrieval_setup}. 
In outline generation, the system consults 1000–1200 papers, with a maximum of 8 sections.
For section drafting, each subsection retrieves up to 60 additional relevant papers, combined with those linked during outline generation.
Finally, we apply two iterations of the review-and-refine loop to enhance coherence across sections and improve overall readability.
Illustrative outputs compared with \as\ are provided in App.~\ref{subsec:Comparison_AutoSurvey_IterSurvey}.

\paragraph{Baselines.}
We compare \ours\ with a set of baselines, ranging from simple retrieval-augmented generation (Naive RAG), which directly drafts from retrieved documents, to more advanced state-of-the-art systems. Specifically, we evaluate against AutoSurvey~\citep{wang2024autosurvey}, the first systematic framework for this task; SurveyForge~\citep{yan-etal-2025-surveyforge}, which combines heuristic outline generation based on the logical structures of human-written surveys with a memory-driven scholar navigation agent for high-quality retrieval; and SurveyGo~\citep{wang2025llm}, which employs the LLM$\times$MapReduce-V2 algorithm to address the long-context challenge. We also compare with SurveyX~\citep{liang2025surveyx}, which introduces an Attribute Tree-based outlining mechanism; however, due to access restrictions, we include SurveyX only in arena experiments. All methods are evaluated on the same retrieval database with generation hyperparameters aligned to their original settings for fairness.

\subsection{Automatic Evaluation Results}
\paragraph{Evaluation Setup.}  
We employ multiple complementary protocols to evaluate the quality of generated surveys. On the 20-topic suite from \citet{wang2024autosurvey}, we adopt multi-dimensional scoring with LLM-as-a-judge. Content quality is assessed along three dimensions: coverage, structure, and relevance followed from \citet{wang2024autosurvey}. Besides, citation quality is evaluated using the NLI-based protocol of \citet{gao2023enabling}, reporting both recall and precision: \textit{Citation Recall} measures whether all statements in the generated text are fully supported by the cited passages, while \textit{Citation Precision} identifies irrelevant citations to ensure that references are pertinent and directly support the claims. To improve scoring stability and reliability, prompts are standardized and judges must provide a rationale before assigning scores. For additional robustness, we aggregate outputs from three judge models: GPT-4o, Claude-3.5-Haiku, and GLM-4.5V.\footnote{Specifically, we use \texttt{chatgpt-4o-latest}, \texttt{claude-3-5-haiku-20241022}, and \texttt{glm-4.5v}.} Full prompts are provided in App.~\ref{app:eval_prompt}.

\paragraph{Results.}
The results on the 20 topics from~\citet{wang2024autosurvey} are reported in Tab.~\ref{tab:main_result}. Statistical significance was confirmed via paired t-tests, indicating that \ours\ consistently outperforms baseline models ($p<0.05$). We summarize the main observations below.
\begin{itemize}
  \item \textbf{Overall superiority.} \ours\ consistently outperforms all baselines across both content and citation quality, achieving the highest overall average score ($4.75$). This demonstrates that the proposed framework is effective and robust across multiple evaluation dimensions.  
  \item \textbf{Improved structural quality.} On the structure dimension, \ours\ achieves the best score ($4.72$). This improvement stems from the recurrent outline generation mechanism, which iteratively explores the literature and refines the outline, resulting in clearer organizational planning and stronger cross-sectional coherence.  
  \item \textbf{Enhanced citation quality.} \ours\ also achieves superior citation performance. While maintaining the same precision as AutoSurvey, it improves recall to $0.70$. This advantage is enabled by paper cards, which provide fine-grained summaries of individual papers and thus allow for retrieving and citing a broader yet still accurate set of supporting references.  
\end{itemize}



Together, these results confirm that recurrent outline generation, paper cards, and outline–paper grounding synergize to produce surveys that are both structurally coherent and rigorously evidenced.

\begin{table}[t]
\centering
\footnotesize
\setlength{\tabcolsep}{3pt}     
\renewcommand{\arraystretch}{1.1}
\caption{Comparison of different methods in terms of content quality and citation quality.}
\label{tab:main_result}
\resizebox{0.90\linewidth}{!}{%
\begin{tabularx}{\linewidth}{l *{4}{>{\centering\arraybackslash}X} *{2}{>{\centering\arraybackslash}X}}
\toprule
\multirow{2}{*}{\textbf{Methods}} &
\multicolumn{4}{c}{\textbf{Content Quality}} &
\multicolumn{2}{c}{\textbf{Citation Quality}} \\
\cmidrule(lr){2-5}\cmidrule(l){6-7}
& \textbf{Coverage} & \textbf{Relevance} & \textbf{Structure} & \textbf{Avg.} & \textbf{Precision} & \textbf{Recall} \\
\midrule
NaiveRAG   & $4.42_{\ \!\pm\!\ 0.50}$ & $4.85_{\ \!\pm\!\ 0.36}$ & $4.20_{\ \!\pm\!\ 0.73}$ & $4.49_{\ \!\pm\!\ 0.41}$ & $0.39_{\ \!\pm\!\ 0.16}$ & $0.40_{\ \!\pm\!\ 0.15}$ \\
\as                 & $4.50_{\ \!\pm\!\ 0.29}$ & $4.80_{\ \!\pm\!\ 0.16}$ & $4.62_{\ \!\pm\!\ 0.24}$ & $4.64_{\ \!\pm\!\ 0.15}$ & $0.64_{\ \!\pm\!\ 0.08}$ & $0.64_{\ \!\pm\!\ 0.08}$ \\
SurveyForge & $4.57_{\ \!\pm\!\ 0.50}$ & $4.82_{\ \!\pm\!\ 0.39}$ & $4.60_{\ \!\pm\!\ 0.56}$ & $4.66_{\ \!\pm\!\ 0.40}$ & $0.59_{\ \!\pm\!\ 0.09}$ & $0.59_{\ \!\pm\!\ 0.09}$ \\
SurveyGo   & $4.37_{\ \!\pm\!\ 0.49}$ & $4.83_{\ \!\pm\!\ 0.38}$ & $4.27_{\ \!\pm\!\ 0.63}$ & $4.49_{\ \!\pm\!\ 0.40}$ & $0.50_{\ \!\pm\!\ 0.11}$ & $0.63_{\ \!\pm\!\ 0.12}$ \\
\rowcolor[HTML]{BCA88D} 
\ours      & \textbf{$\textbf{4.58}_{\ \!\pm\!\ 0.50}$} & \textbf{$\textbf{4.95}_{\ \!\pm\!\ 0.22}$} & \textbf{$\textbf{4.72}_{\ \!\pm\!\ 0.45}$} & \textbf{$\textbf{4.75}_{\ \!\pm\!\ 0.30}$} & \textbf{$\textbf{0.64}_{\ \!\pm\!\ 0.06}$} & \textbf{$\textbf{0.70}_{\ \!\pm\!\ 0.07}$} \\
\bottomrule
\end{tabularx}
}

\end{table}

\subsection{Human Evaluation Results}
\begin{figure}[t]
  \centering
  \begin{subfigure}[t]{0.49\linewidth}
    \centering
    \includegraphics[width=\linewidth]{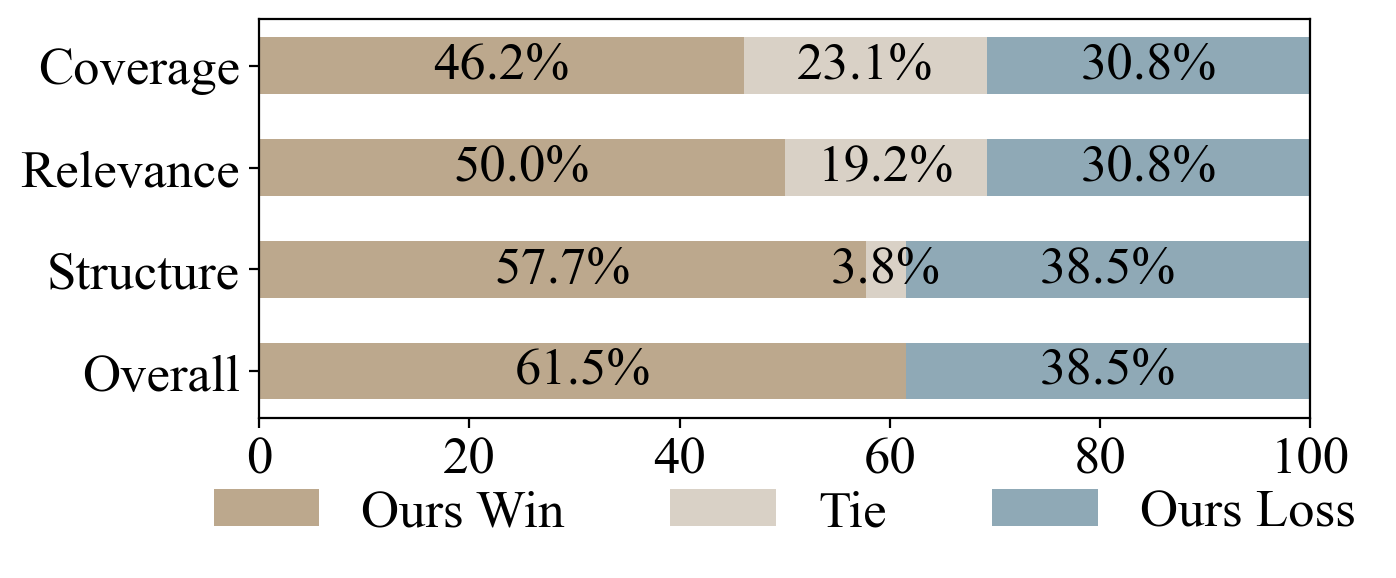}
    \caption{IterSurvey vs AutoSurvey}
    \label{fig:iter_vs_auto}
  \end{subfigure}
  \hfill
  \begin{subfigure}[t]{0.49\linewidth}
    \centering
    \includegraphics[width=\linewidth]{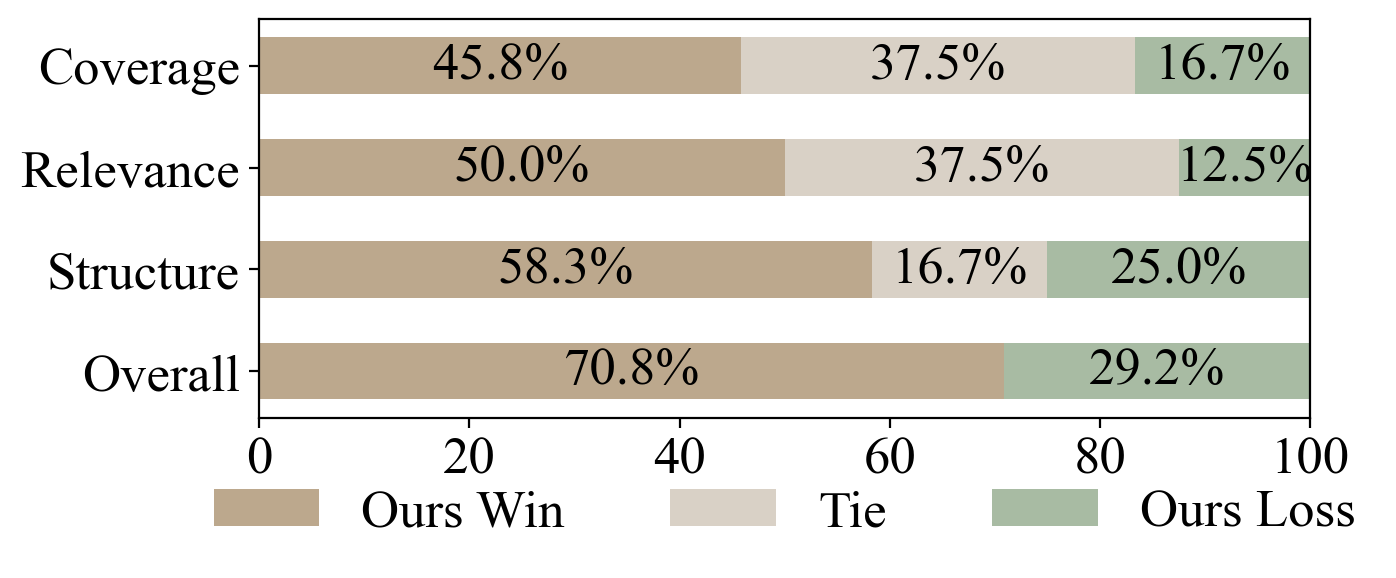}
    \caption{IterSurvey vs SurveyForge}
    \label{fig:iter_vs_forge}
  \end{subfigure}
  \caption{LLM-generated survey comparison between AutoSurvey and IterSurvey. }
  \label{fig:human_evaluation}
  \vspace{-1em}
\end{figure}

To further assess the quality of the generated surveys, we conducted a blind, pairwise study~\citep{novikova2018rankme,chiang2024chatbotarenaopenplatform} with seven PhD-level experts. 
For each evaluation, experts were presented with an anonymized survey pair and asked to select the superior one based on multiple quality dimensions, including coverage, relevance, structural coherence, and overall quality, which is more objective and stable than ranking based on absolute scores~\citep{herbrich2006trueskill, sakaguchi-etal-2014-efficient}. To control annotation cost, the human study was limited to direct comparisons between \ours\ and two leading baselines: AutoSurvey and SurveyForge. Inter-rater agreement is reported in App.~\ref{app:inter_rater_agreement}.
Results, as shown in Fig.~\ref{fig:human_evaluation}, indicate that \ours\ is consistently preferred over AutoSurvey and SurveyForge by domain experts, especially in terms of structure and overall quality. 
This trend aligns with our automatic evaluation, where recurrent outline generation also demonstrated stronger coherence and organization. 
The consistency between expert judgments and automatic metrics further highlights the robustness of \ours\ in generating high-quality surveys.

\subsection{Survey-Arena: Pairwise Comparison and Ranking}
\label{sec:survey-arena}

\paragraph{Dataset construction.}  
Previous automatic evaluation methods typically assign an absolute score for each dimension, which struggles to fully capture the performance gap between machine-generated surveys and human-written ones. To move beyond absolute scores, we constructed the \textit{Survey-Arena} benchmark. The benchmark spans ten research topics. For each topic, we manually selected five high-quality, human-written surveys to serve as a performance baseline. To ensure comparability, all surveys for a given topic were chosen from a narrow six-month submission window, a process that required careful verification to ensure each topic had a sufficient number of suitable papers. We further confirmed their quality and influence via non-trivial citation counts on Google Scholar. The retrieval database for all machine-generated surveys was correspondingly frozen to the same time period to guarantee fairness. The full list of topics and papers is available in the App.~\ref{app:arena_topic}.

\paragraph{Evaluation protocol.}  
For each topic, all possible pairs of a machine-generated survey and a human-written survey are constructed. To ensure robust evaluation and mitigate positional bias, each pair is judged in both directions (A vs. B and B vs. A), following \cite{li2024splitmergealigningposition}. A panel of three distinct LLMs, namely GPT-4o, Claude-3.5-Haiku, and GLM-4.5V, serves as the judges for each comparison. Elo scores are computed from these aggregated pairwise outcomes to generate rankings for all systems.

\paragraph{Results.}

\begin{wrapfigure}{r}{0.48\textwidth}
    \centering
    \vspace{-15pt}
    \begin{minipage}{\linewidth}

        \includegraphics[width=\linewidth]{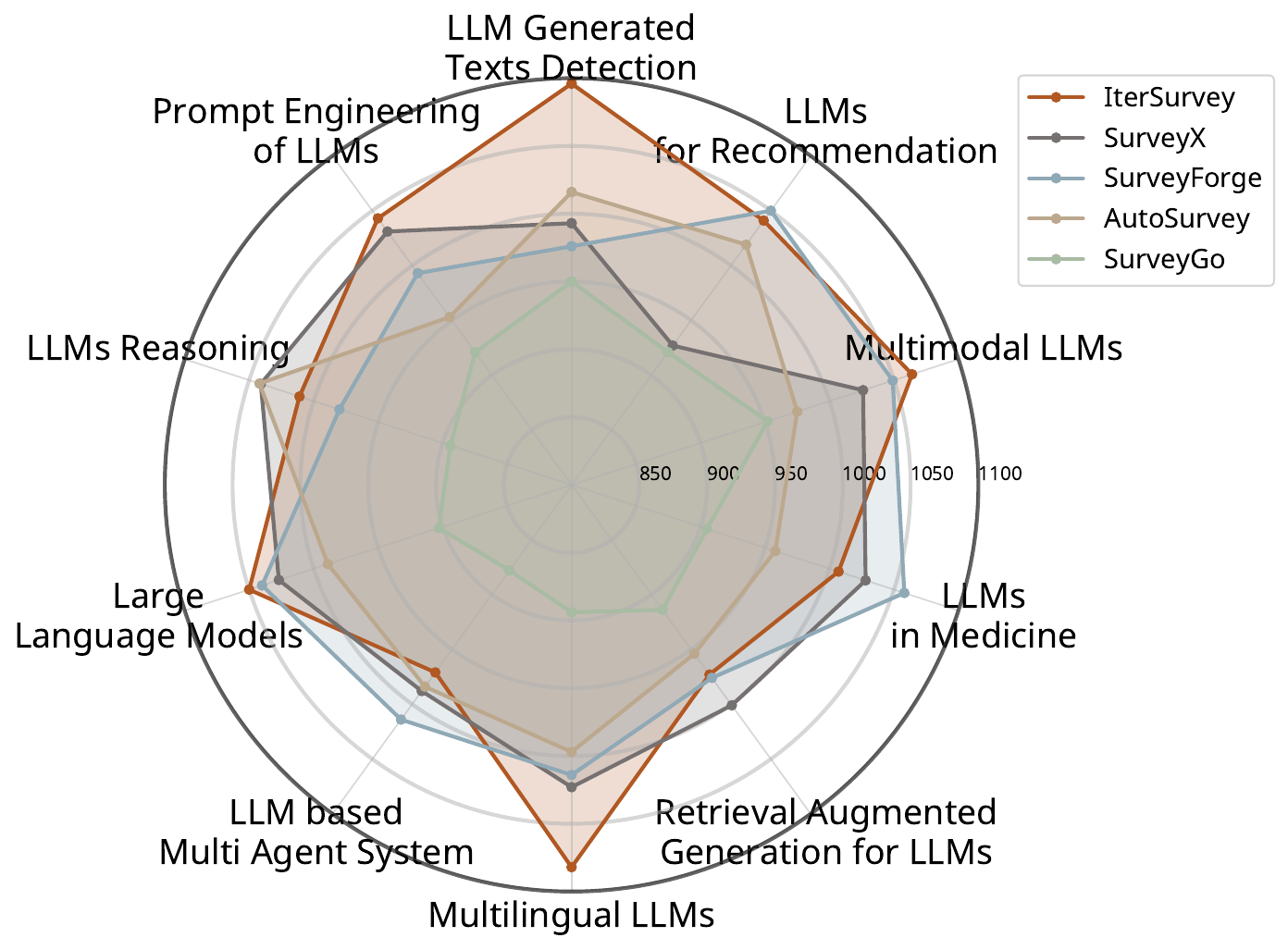}
        \captionof{figure}{Elo scores of Survey-Arena results across topics. The radar plot shows the Elo scores for each system across all topics, providing a topic-wise comparison.}
        \label{fig:arena_result}

        \captionof{table}{Aggregated rankings on Survey-Arena. Avg. Rank is the mean position among all surveys. $>$Human\% is the average proportion of topics where a system surpasses human surveys.}
        \label{tab:arena_rankings}
        \vspace{6pt}
        \footnotesize
        \setlength{\tabcolsep}{4pt}
        \renewcommand{\arraystretch}{1.15}
        \begin{tabularx}{\linewidth}{l >{\centering\arraybackslash}X >{\centering\arraybackslash}X}
            \toprule
            \textbf{Method} & \textbf{Avg. Rank} $\downarrow$ & \textbf{$>$ Human\%} $\uparrow$ \\
            \midrule
            SurveyGo     & 9.80 &  4\% \\
            AutoSurvey   & 6.70 & 32\% \\
            SurveyForge  & 4.80 & 50\% \\
            SurveyX      & 4.70 & 54\% \\
            \rowcolor[HTML]{BCA88D}
            \textbf{\ours} & \textbf{4.00} & \textbf{60\%} \\
            \bottomrule
        \end{tabularx}
        \vspace{6pt}
        \captionof{table}{Consistency between different ranking methods and citation-based rankings.}
        \label{tab:arena_meta_eval}
        \vspace{6pt}
        \footnotesize
        \setlength{\tabcolsep}{4pt}
        \renewcommand{\arraystretch}{1.15}
        \begin{tabularx}{\linewidth}{l >{\centering\arraybackslash}X >{\centering\arraybackslash}X >{\centering\arraybackslash}X}
            \toprule
            \textbf{Rank Method} & $\boldsymbol{\rho}_{\mathrm{s}}$ & \textbf{nDCG@2} & \textbf{nDCG@3} \\
            \midrule
            Absolute Scoring & 0.320 & 0.695 & 0.767 \\
            Pair-Judge       & \textbf{0.410} & \textbf{0.834} & \textbf{0.873} \\
            \bottomrule
        \end{tabularx}

    \end{minipage}
    \vspace{-30pt}
\end{wrapfigure}

We report two key evaluation metrics: Avg. Rank, which indicates the mean position among all surveys, and $>$Human\%, which reflects the proportion of topics where a system surpasses human surveys. The topic-wise outcomes from Survey-Arena are visualized in Fig.\ref{fig:arena_result}, and the aggregated rankings are summarized in Tab.\ref{tab:arena_rankings}.

Each system is evaluated by its average rank across all surveys (including 5 machine-written surveys and 5 human-written ones) and by the proportion of topics where it surpasses human surveys.
The results show that \ours\ consistently achieves the best overall performance among automatic survey generation systems, with an average rank of $4.0$ and surpassing human-written surveys in $60\%$ of topics.
These findings highlight that \ours\ not only outperforms competing methods but also approaches human-level quality across diverse domains.

\paragraph{Meta Evaluation.} 
To assess the reliability of Survey-Arena judgments, we compare the rankings produced by Survey-Arena for human-written surveys with citation counts on Google Scholar, which serve as an external signal of impact. Specifically, we compute Spearman’s $\rho_s$ by measuring the correlation between Arena-derived and citation-based rankings for each topic, and then report the average across topics. For relevance scoring, we treat citation counts as an indicator of relevance and compute nDCG directly over the ranking lists. As a comparison, we also use the rankings derived from absolute scoring and compute their consistency and nDCG. This allows us to evaluate how well the different ranking methods align with citation-based rankings.

Results are shown in Tab.~\ref{tab:arena_meta_eval}. Compared with the scoring-based approach, pairwise judgment achieves higher agreement with citation-based rankings, yielding a Spearman’s $\rho_s$ of $0.410$ and nDCG@$2/3 =$ $0.834/0.873$. This indicates that when models are asked to directly compare two surveys, they more reliably identify the superior one, producing rankings that better align with human impact signals. These findings support pairwise evaluation as a more robust protocol for Survey-Arena.


\subsection{Generalization on Survey-Lacking Topics}
To examine whether automated survey generation can succeed in areas without existing surveys, we construct a subset of eight research topics (listed in App.~\ref{app:ood_topic}) where no human-written reviews are available. Such settings are common in emerging domains and pose greater challenges, since there are no canonical structures to imitate and the literature is often sparse and fragmented. This setup tests whether a system can autonomously organize the field into a coherent, well-grounded survey.

We compare \ours\ against AutoSurvey and SurveyForge under this setup, and the results are presented in Tab.~\ref{tab:ood_result}.
Our method achieves the highest average score ($4.63$), consistently outperforming both baselines across content and citation quality.
Notably, \ours\ shows clear advantages in structural quality ($4.63$) and citation recall ($0.67$).
These gains highlight the benefits of recurrent outline generation, which encourages iterative query expansion and literature exploration rather than relying on a fixed set of initial retrievals.
Combined with paper cards providing fine-grained evidence abstraction, this mechanism enables \ours\ to construct coherent survey structures and incorporate broader supporting references even in areas where survey conventions are absent.
\begin{table}[t]
\centering
\footnotesize
\setlength{\tabcolsep}{3pt}     
\renewcommand{\arraystretch}{1.1}
\caption{Comparison of different methods on survey-lacking topics.}
\label{tab:ood_result}
\resizebox{0.90\linewidth}{!}{%
\begin{tabularx}{\linewidth}{l *{4}{>{\centering\arraybackslash}X} *{2}{>{\centering\arraybackslash}X}}
\toprule
\multirow{2}{*}{\textbf{Methods}} &
\multicolumn{4}{c}{\textbf{Content Quality}} &
\multicolumn{2}{c}{\textbf{Citation Quality}} \\
\cmidrule(lr){2-5}\cmidrule(l){6-7}
& \textbf{Coverage} & \textbf{Relevance} & \textbf{Structure} & \textbf{Avg.} & \textbf{Precision} & \textbf{Recall} \\
\midrule
\as                 & $4.00_{\ \!\pm\!\ 1.12}$ & $4.20_{\ \!\pm\!\ 1.20}$ & $4.00_{\ \!\pm\!\ 1.00}$ & $4.07_{\ \!\pm\!\ 1.11}$ & $0.55_{\ \!\pm\!\ 0.14}$ & $0.55_{\ \!\pm\!\ 0.09}$ \\
SurveyForge& $\textbf{4.50}_{\ \!\pm\!\ 0.50}$ & $4.75_{\ \!\pm\!\ 0.50}$ & $4.54_{\ \!\pm\!\ 0.54}$ & $4.60_{\ \!\pm\!\ 0.52}$ & $0.47_{\ \!\pm\!\ 0.12}$ & $0.47_{\ \!\pm\!\ 0.13}$ \\
\rowcolor[HTML]{BCA88D}
\ours      & \textbf{$4.42_{\ \!\pm\!\ 0.58}$} & \textbf{$\textbf{4.83}_{\ \!\pm\!\ 0.17}$} & \textbf{$\textbf{4.63}_{\ \!\pm\!\ 0.63}$} & \textbf{$\textbf{4.63}_{\ \!\pm\!\ 0.37}$} & \textbf{$\textbf{0.60}_{\ \!\pm\!\ 0.06}$} & \textbf{$\textbf{0.67}_{\ \!\pm\!\ 0.06}$} \\
\bottomrule
\end{tabularx}
}

\end{table}
\subsection{Ablation Study}
\begin{table}[t]
\centering
\footnotesize
\setlength{\tabcolsep}{3pt}     
\renewcommand{\arraystretch}{1.1}
\caption{Ablation study analyzing the contribution of each component in \ours: $\Circle$ Recurrent outline generation; $\Square$ Paper Card; $\Diamond$ Review-and-Refine.}
\label{tab:ablation}
\resizebox{0.90\linewidth}{!}{
\begin{tabularx}{\linewidth}{l *{4}{>{\centering\arraybackslash}X} *{2}{>{\centering\arraybackslash}X}}
\toprule
\multirow{2}{*}{\textbf{Methods}} &
\multicolumn{4}{c}{\textbf{Content Quality}} &
\multicolumn{2}{c}{\textbf{Citation Quality}} \\
\cmidrule(lr){2-5}\cmidrule(l){6-7}
& \textbf{Coverage} & \textbf{Relevance} & \textbf{Structure} & \textbf{Avg.} & \textbf{Precision} & \textbf{Recall} \\
\midrule
Baseline                                & \(4.00_{\ \!\pm\!\ 0.53}\) & \(4.40_{\ \!\pm\!\ 0.48}\) & \(4.20_{\ \!\pm\!\ 0.70}\) & \(4.20_{\ \!\pm\!\ 0.44}\) & \(0.58_{\ \!\pm\!\ 0.09}\) & \(0.67_{\ \!\pm\!\ 0.09}\) \\
+ \(\Circle\)                           & \(4.46_{\ \!\pm\!\ 0.52}\) & \(4.80_{\ \!\pm\!\ 0.41}\) & \(4.53_{\ \!\pm\!\ 0.52}\) & \(4.60_{\ \!\pm\!\ 0.40}\) & \(0.62_{\ \!\pm\!\ 0.08}\) & \(0.59_{\ \!\pm\!\ 0.09}\) \\
+ \(\Circle\)  +\ \(\Square\)           & \(4.60_{\ \!\pm\!\ 0.51}\) & \(4.80_{\ \!\pm\!\ 0.42}\) & \(4.60_{\ \!\pm\!\ 0.52}\) & \(4.69_{\ \!\pm\!\ 0.39}\) & \(0.64_{\ \!\pm\!\ 0.09}\) & \(0.71_{\ \!\pm\!\ 0.08}\) \\
+ \(\Circle\) + \(\Square\) +\ \(\Diamond\)  & \(4.73_{\ \!\pm\!\ 0.50}\) & \(4.93_{\ \!\pm\!\ 0.41}\) & \(4.80_{\ \!\pm\!\ 0.52}\) & \(4.82_{\ \!\pm\!\ 0.39}\) & \(0.65_{\ \!\pm\!\ 0.04}\) & \(0.77_{\ \!\pm\!\ 0.04}\) \\
\bottomrule
\end{tabularx}
}
\vspace{-10px}
\end{table}
We conducted an ablation study on five representative topics to analyze the impact of the three new modules of \ours: Recurrent Outline Generation, Paper Card, and Review-and-Refine. Results are shown in Tab.~\ref{tab:ablation}, revealing the following insights:

\textbf{Recurrent Outline Generation yields stronger content quality.}
We compare our recurrent outline generation with a one-shot paradigm, where retrieved papers are partitioned into groups, each group produces an outline independently, and the results are subsequently merged. The recurrent approach contributes significant improvements in content quality, with gains of $+0.46$ in coverage and $+0.33$ in structure over the baseline. This demonstrates that iterative exploration helps the model achieve broader coverage and stronger organizational coherence by progressively integrating evidence.

\textbf{Paper Card improves citation quality.}
We further examined the impact of paper card, we replace them with abstract-based inputs commonly used in retrieval pipelines. The results show that paper cards significantly improve citation grounding, raising recall from $0.59$ to $0.71$ while maintaining precision ($0.64$). This indicates that distilled paper-level evidence reduces distraction and enables the model to retrieve and cite a broader yet accurate set of references.

\textbf{Review and Refine boosts overall performance.}
Finally, we evaluate the review-and-refine stage by removing it from the pipeline. The full variant enhances all dimensions of content quality, raising the overall average from $4.69$ to $4.82$, and further improves citation recall from $0.71$ to $0.77$. These gains show that multi-round self-critique and revision help fill evidence gaps, eliminate unsupported claims, and polish the text into well-substantiated surveys. Together, recurrent planning, paper cards, and review-and-refine form the most effective configuration of \ours.

\section{Conclusion}
In this work, we tackled the limitations of existing survey generation systems by introducing \ours, a framework with recurrent outline generation, paper cards, and global review and integration. This design enables precise retrieval, coherent structure, and faithful citation grounding, while supporting multimodal outputs. Experiments on diverse topics show that \ours\ outperforms state-of-the-art baselines in coherence, coverage, and citation quality. We also proposed Survey-Arena, a pairwise benchmark that complements absolute scoring for a more reliable assessment. Future work will extend our framework to broader domains, integrate richer multimodal evidence, and refine evaluation protocols toward human-level quality.

\section*{Ethical Considerations}
Our work focuses on automatic literature survey generation using large language models. While the system is designed to support researchers by synthesizing existing knowledge, it inevitably inherits limitations of current models, including potential citation errors, incomplete coverage, and occasional inaccuracies. Therefore, the generated surveys are intended as an assistive tool rather than a substitute for human scholarship, and should be used for reference only. For evaluation, all human experts involved in the study participated voluntarily and received fair compensation. All data used in our experiments were sourced from publicly available arXiv papers, which permit non-commercial use. We strictly avoided the use of private or sensitive data.

\section*{Use of Large Language Models}
We used large language models (GPT-4o, Claude-3.5-Haiku, and GLM-4.5V) in two ways: 
(i) as evaluation judges for assessing survey quality, and 
(ii) for limited language editing and refinement of the manuscript. 
All substantive research ideas, experimental design, analyses, and final decisions were made solely by the authors, 
who take full responsibility for the content of this paper.

\bibliography{iclr2026_conference}
\bibliographystyle{iclr2026_conference}

\newpage
\appendix
\section{Appendix}
\subsection{Retrieval Setup}
\label{app:retrieval_setup}
For the retrieval process, we implemented a lightweight database to provide the necessary functionality. The retrieval logic is based on vector similarity, using the \textit{nomic-ai/nomic-embed-text-v1.5}~\citep{nussbaum2024nomic} embedding model with all hyperparameters set to their default values. Given a query, the database computes the similarity between the query vector and all paper vectors, and returns the top-k most relevant entries. In addition, the database supports bidirectional lookup between a paper’s arXiv identifier and title, as well as filtering papers published prior to a specified cutoff date.

\subsection{Results of Inter-rater Agreement}
To assess the reliability of human annotations, we computed Cohen’s kappa coefficients across four evaluation dimensions: Coverage, Relevance, Structure, and Overall, as shown in Tab.~\ref{tab:app_kappa}. These results indicate substantial agreement among human annotators, supporting the consistency of the human evaluation process.
\label{app:inter_rater_agreement}
\begin{table}[h!]
\centering
\caption{Inter-rater agreement among human annotators.}
\label{tab:app_kappa}
\begin{tabular}{@{}ccccc@{}}
\toprule
Dimensions & Coverage & Relevance & Structure & Overall \\ \midrule
kappa      & 0.714    & 0.583     & 0.611     & 0.650   \\ \bottomrule
\end{tabular}
\end{table}

\subsection{Topics for Automatic Evaluation}
\label{app:automatic_topic}
We utilize 20 topics derived from AutoSurvey~\citep{wang2024autosurvey}. Each topic is paired with a human survey, as shown in Tab.~\ref{tab:app_as_topics}, which also reports the survey titles, arXiv IDs, and their latest citation counts from Google Scholar.
\begin{table}[h!]
\centering
\tiny
\caption{Topics for Automatic Evaluation}
\label{tab:app_as_topics}
\begin{tabular}{@{}llrr@{}}
\toprule
\multicolumn{1}{c}{\textbf{Topic}}                     & \multicolumn{1}{c}{\textbf{Human Survey}}                                                          & \multicolumn{1}{c}{\textbf{ArXiv ID}} & \textbf{Citations} \\ \midrule
In-context Learning               & A Survey on In-context Learning                                                & 2301.00234        & 2396               \\
LLMs for Recommendation           & A Survey on Large Language Models for Recommendation                           & 2305.19860        & 596                \\
LLM-Generated Texts Detection     & The Science of Detecting LLM-Generated Texts                                   & 2310.14724        & 308                \\
Explainability for LLMs           & Explainability for Large Language Models: A Survey                             & 2309.01029        & 875                \\
Evaluation of LLMs                & A Survey on Evaluation of Large Language Models                                & 2307.03109        & 4020               \\
LLMs-based Agents                 & A Survey on Large Language Model based Autonomous Agents                       & 2308.11432        & 1906               \\
LLMs in Medicine                  & A Survey of Large Language Models in Medicine                                  & 2311.05112        & 217                \\
Domain Specialization of LLMs     & Domain Specialization as the Key to Make Large Language Models Disruptive      & 2305.18703        & 217                \\
Challenges of LLMs in Education   & Practical and Ethical Challenges of Large Language Models in Education         & 2303.13379        & 722                \\
Alignment of LLMs                 & Aligning Large Language Models with Human: A Survey                            & 2307.12966        & 435                \\
ChatGPT                           & Harnessing the Power of LLMs in Practice: A Survey on ChatGPT and Beyond       & 2304.13712        & 1254               \\
Instruction Tuning for LLMs       & Instruction Tuning for Large Language Models: A Survey                         & 2308.10792        & 1174               \\
LLMs for Information Retrieval    & Large Language Models for Information Retrieval: A Survey                      & 2308.07107        & 544                \\
Safety in LLMs                    & Towards Safer Generative Language Models                                       & 2302.09270        & 13                 \\
Chain of Thought                  & A Survey of Chain of Thought Reasoning: Advances, Frontiers and Future         & 2309.15402        & 290                \\
Hallucination in LLMs             & A Survey on Hallucination in Large Language Models                             & 2311.05232        & 2599               \\
Bias and Fairness in LLMs         & Bias and Fairness in Large Language Models: A Survey                           & 2309.00770        & 1009               \\
Large Multi-Modal Language Models & Large-scale Multi-Modal Pre-trained Models: A Comprehensive Survey             & 2302.10035        & 285                \\
Acceleration for LLMs             & A Survey on Model Compression and Acceleration for Pretrained Language Models  & 2202.07105        & 101                \\
LLMs for Software Engineering     & Large Language Models for Software Engineering: A Systematic Literature Review & 2308.10620        & 1058               \\ \bottomrule
\end{tabular}

\end{table}

\newpage
\subsection{Topics for Survey-Arena}
\label{app:arena_topic}
To construct the Survey-Arena benchmark, we select 10 topics, with several derived from AutoSurvey~\citep{wang2024autosurvey} and SurveyForge~\citep{yan-etal-2025-surveyforge}. For each topic, we include 5 human-written surveys, requiring that their arXiv submission dates fall within a six-month window. We report their latest Google Scholar citation counts as a measure of impact, as summarized in Tab.~\ref{tab:app_arena_topics}. For reproducibility, we also specify the exact arXiv version, since submission dates can vary considerably across different versions of the same paper.
\begin{table}[h!]
\footnotesize
\centering
\tiny
\caption{Topics for Survey-Arena}
\label{tab:app_arena_topics}
\begin{tabular}{@{}llrr@{}}
\toprule
\multicolumn{1}{c}{\textbf{Topic}}                                                                  & \multicolumn{1}{c}{\textbf{Human Survey}}                                                                                                  & \multicolumn{1}{c}{\textbf{ArXiv ID}} & \textbf{Citations} \\ \midrule
\multirow{5}{*}{\begin{tabular}[c]{@{}l@{}}Large Language \\ Models\end{tabular}}                   & Large Language Models: A Survey                                                                                                            & 2402.06196v3                          & 1133               \\
                                                                                                    & Large Language Models Meet NLP: A Survey                                                                                                   & 2405.12819v1                          & 86                 \\
                                                                                                    & History, Development, and Principles of Large Language Models-An Introductory Survey                                                       & 2402.06853v2                          & 73                 \\
                                                                                                    & Recent Advances in Generative AI and Large Language Models                                                                                 & 2407.14962v1                          & 68                 \\
                                                                                                    & Exploring the landscape of large language models: Foundations, techniques, and challenges                                                  & 2404.11973v1                          & 5                  \\ \midrule
\multirow{5}{*}{Multimodal LLMs}                                                                    & MM-LLMs: Recent Advances in MultiModal Large Language Models                                                                               & 2401.13601v3                          & 381                \\
                                                                                                    & Multimodal Large Language Models: A Survey                                                                                                 & 2311.13165v1                          & 299                \\
                                                                                                    & The Revolution of Multimodal Large Language Models: A Survey                                                                               & 2402.12451v1                          & 98                 \\
                                                                                                    & How to Bridge the Gap between Modalities: Survey on Multimodal Large Language Model                                                        & 2311.07594v1                          & 43                 \\
                                                                                                    & A Review of Multi-Modal Large Language and Vision Models                                                                                   & 2404.01322v1                          & 39                 \\ \midrule
\multirow{5}{*}{Multilingual LLMs}                                                                  & Multilingual Large Language Model: A Survey of Resources, Taxonomy and Frontiers                                                           & 2404.04925v1                          & 83                 \\
                                                                                                    & A Survey on Multilingual Large Language Models: Corpora, Alignment, and Bias                                                               & 2404.00929v2                          & 55                 \\
                                                                                                    & A Survey on Large Language Models with Multilingualism                                                                                     & 2405.10936v1                          & 40                 \\
                                                                                                    & Surveying the MLLM Landscape: A Meta-Review of Current Surveys                                                                             & 2409.18991v1                          & 12                 \\
                                                                                                    & Multilingual Large Language Models: A Systematic Survey                                                                                    & 2411.11072v2                          & 9                  \\ \midrule
\multirow{5}{*}{LLMs Reasoning}                                                                     & A Survey of Long Chain-of-Thought for Reasoning Large Language Models                                                                      & 2503.09567v3                          & 130                \\
                                                                                                    & From System 1 to System 2: A Survey of Reasoning Large Language Models                                                                     & 2502.17419v2                          & 110                \\
                                                                                                    & Advancing Reasoning in Large Language Models: Promising Methods and Approaches                                                             & 2502.03671v1                          & 19                 \\
                                                                                                    & A Survey of Frontiers in LLM Reasoning                                                                                                     & 2504.09037v1                          & 17                 \\
                                                                                                    & Thinking Machines: A Survey of LLM based Reasoning Strategies                                                                              & 2503.10814v1                          & 9                  \\ \midrule
\multirow{5}{*}{\begin{tabular}[c]{@{}l@{}}Prompt Engineering\\ of LLMs\end{tabular}}               & A Systematic Survey of Prompt Engineering in Large Language Models                                                                         & 2402.07927v1                          & 748                \\
                                                                                                    & The Prompt Report: A Systematic Survey of Prompt Engineering Techniques                                                                    & 2406.06608v2                          & 182                \\
                                                                                                    & Prompt Design and Engineering: Introduction and Advanced Methods                                                                           & 2401.14423v4                          & 117                \\
                                                                                                    & A Survey of Prompt Engineering Methods in Large Language Models for Different NLP Tasks                                                    & 2407.12994v1                          & 60                 \\
                                                                                                    & Efficient Prom pting Methods for Large Language Models: A Survey                                                                           & 2404.01077v1                          & 56                 \\ \midrule
\multirow{5}{*}{\begin{tabular}[c]{@{}l@{}}Retrieval-Augmented \\ Generation for LLMs\end{tabular}} & Retrieval-Augmented Generation for Large Language Models: A Survey                                                                         & 2312.10997v5                          & 2583               \\
                                                                                                    & A Survey on RAG Meeting LLMs: Towards Retrieval-Augmented Large Language Models                                                            & 2405.06211v3                          & 559                \\
                                                                                                    & A Survey on Retrieval-Augmented Text Generation for Large Language Models                                                                  & 2404.10981v2                          & 119                \\
                                                                                                    & Retrieval-Augmented Generation for Natural Language Processing: A Survey                                                                   & 2407.13193v2                          & 77                 \\
                                                                                                    & Retrieval Augmented Generation (RAG) and Beyond                                                                                            & 2409.14924v1                          & 70                 \\ \midrule
\multirow{5}{*}{\begin{tabular}[c]{@{}l@{}}LLM-based\\ Multi-Agent System\end{tabular}}             & A survey on large language model based autonomous agents                                                                                   & 2308.11432v7                          & 1623               \\
                                                                                                    & Multi-Agent Collaboration Mechanisms: A Survey of LLMs                                                                                     & 2501.06322v1                          & 79                 \\
                                                                                                    & Large language model agent: A survey on methodology, applications and challenges                                                           & 2503.21460v1                          & 19                 \\
                                                                                                    & Agentic large language models, a survey                                                                                                    & 2503.23037v2                          & 12                 \\
                                                                                                    & A Survey on LLM-based Multi-Agent System:                                                                                                  & 2412.17481v2                          & 3                  \\ \midrule
\multirow{5}{*}{\begin{tabular}[c]{@{}l@{}}LLM-Generated\\ Texts Detection\end{tabular}}            & A Survey on LLM-Generated Text Detection: Necessity, Methods, and Future Directions                                                        & 2310.14724v2                          & 210                \\
                                                                                                    & A Survey on Detection of LLMs-Generated Content                                                                                            & 2310.15654v1                          & 69                 \\
                                                                                                    & Towards Possibilities \& Impossibilities of AI-generated Text Detection: A Survey                                                          & 2310.15264v1                          & 46                 \\
                                                                                                    & Detecting chatgpt: A survey of the state of detecting chatgpt-generated text                                                               & 2309.07689v1                          & 22                 \\
                                                                                                    & Decoding the AI Pen: Techniques and Challenges in Detecting AI-Generated Text                                                              & 2403.05750v1                          & 13                 \\ \midrule
\multirow{5}{*}{LLMs in Medicine}                                                                   & Large language models in healthcare and medical domain: A review                                                                           & 2401.06775v2                          & 246                \\
                                                                                                    & A Survey on Medical Large Language Models                                                                                                  & 2406.03712v1                          & 53                 \\
                                                                                                    & \begin{tabular}[c]{@{}l@{}}A Comprehensive Survey of Large Language Models and Multimodal Large Language Models\\ in Medicine\end{tabular} & 2405.08603v1                          & 46                 \\
                                                                                                    & Large Language Models for Medicine: A Survey                                                                                               & 2405.13055v1                          & 37                 \\
                                                                                                    & A Comprehensive Survey on Evaluating Large Language Model Applications in the Medical Industry                                             & 2404.15777v4                          & 32                 \\ \midrule
\multirow{5}{*}{\begin{tabular}[c]{@{}l@{}}LLMs for\\ Recommendation\end{tabular}}                  & A Survey on Large Language Models for Recommendation                                                                                       & 2305.19860v4                          & 508                \\
                                                                                                    & Recommender Systems in the Era of Large Language Models (LLMs)                                                                             & 2307.02046v2                          & 479                \\
                                                                                                    & A Comprehensive Survey of Language Modelling Paradigm Adaptations in Recommender Systems                                                   & 2302.03735v3                          & 117                \\
                                                                                                    & Large Language Models for Generative Recommendation: A Survey and Visionary Discussions                                                    & 2309.01157v1                          & 116                \\
                                                                                                    & How Can Recommender Systems Benefit from Large Language Models: A Survey                                                                   & 2306.05817v4                          & 104                \\ \bottomrule
\end{tabular}

\end{table}

\newpage
\subsection{Topics for Survey-Lacking Test}
\label{app:ood_topic}
We manually select 8 topics with no existing survey articles, as shown in Tab.~\ref{tab:ood_topic}.
\begin{table}[h!]
\centering
\tiny
\caption{Topics for Survey-Lacking Test.}
\label{tab:ood_topic}
\begin{tabular}{@{}c@{}}
\toprule
\textbf{Topic}                                                                    \\ \midrule
Event Timeline Generation                                                \\
Linear RNN in Natural Language Processing                                \\
Agent-flow Data Curation                                                 \\
Causal Mediation with Sparse Autoencoder Features in Transformers        \\
Multi-Tenant Scheduling for MoE Inference                                \\
Benchmarking Tool-Using LLMs for Causal Tasks in the MCP Ecosystem       \\
RAG for Mechanical Design: Cross-Modal Retrieval over CAD Trees and BOMs \\
Renderer-in-the-Loop Supervision for Multimodal Model                    \\ \bottomrule
\end{tabular}

\end{table}

\subsection{Detail of Naive RAG}
Given a topic, the Naive RAG system first retrieves 1,500 papers from the same database as ours. It then employs an iterative prompting strategy, where the LLM generates content until the total length of the survey reaches 5,000 tokens~\citep{wang2024autosurvey}. The prompt used for generation is shown below.
\label{app:naive_prompt}
\begin{tcblisting}{
  listing only,
  listing engine=listings,  % 原样显示，不转义
  enhanced, breakable,      % 可跨页
  colback=blue!6, colframe=blue!60!black,
  arc=3mm, boxrule=0.6pt,,
  left=6mm, right=6mm, top=4mm, bottom=4mm,
  title={Naive RAG Prompt}
}
You are an expert in artificial intelligence who wants to write an overall and comprehensive survey about [TOPIC].

You are provided with a list of papers related to [TOPIC] below:
---
[PAPER LIST]
---

Here is the survey content you have written:
---
[SURVEY CONTENT]
---

Hers is the requirement of the survey:
1. The survey must be more than [SURVEY LEN] tokens!
2. Containing serval sections. Each section contains several subsections.
3. Cite several paper provided above to support the content you write.

Here is the format of your writing:
1. ’##’ indicates the section title
2. ’###’ indicates the subsection title
3. Only cite the "paper_title" in ’[]’. An example of citation: ’the emergence of large language models (LLMs) [Language models are few-shot learners; Language models are unsupervised multitask learners; PaLM: Scaling language modeling with pathways]’

You need to continue writing the survey by adding a new section or subsection.

Do not stop until the length of survey is more than [SURVEY LEN] tokens!!!

Return the content you write:
\end{tcblisting}

\subsection{Prompts for Evaluation}
\label{app:eval_prompt}
\begin{tcblisting}{
  listing only,
  listing engine=listings,  % 原样显示，不转义
  enhanced, breakable,      % 可跨页
  colback=blue!6, colframe=blue!60!black,
  arc=3mm, boxrule=0.6pt,,
  left=6mm, right=6mm, top=4mm, bottom=4mm,
  title={NLI Prompt}
}
---
Claim:
[CLAIM]
---
Source:
[SOURCE]
---
Claim:
[CLAIM]
---
Is the Claim faithful to the Source?
A Claim is faithful to the Source if the core part in the Claim can be supported by the Source.\n
Only reply with 'Yes' or 'No':
\end{tcblisting}

\begin{tcblisting}{
  listing only,
  listing engine=listings,  % 原样显示，不转义
  enhanced, breakable,      % 可跨页
  colback=blue!6, colframe=blue!60!black,
  arc=3mm, boxrule=0.6pt,,
  left=6mm, right=6mm, top=4mm, bottom=4mm,
  title={Criteria-based judging survey prompt}
}
You are an expert academic evaluator specializing in rigorous assessment of academic survey quality. Your task is to conduct a comprehensive evaluation using established scholarly standards and provide detailed justification for your assessment.

<topic>
[TOPIC]
</topic>

<survey_content>
[SURVEY]
</survey_content>

<instruction>
You are provided with:
1. A research topic for context
2. An academic survey for evaluation

Your task is to assess the survey quality based on the specific criterion provided below. Apply rigorous academic standards and provide detailed justification for your assessment. Base your evaluation on specific evidence from the survey content, considering both strengths and areas for improvement.
</instruction>

<evaluation_criterion>
Criterion Description: [Criterion Description]

**CRITICAL: Evaluation Standards**
Your evaluation must follow a systematic approach:

1. **Comprehensive Analysis**: Thoroughly examine the survey content against the specific criterion
2. **Evidence-Based Scoring**: Base your score on specific observable strengths and weaknesses
3. **Detailed Justification**: Provide specific examples and reasoning for your score

**Scoring Framework**:
Score 1: [Score 1 Description]
Score 2: [Score 2 Description]
Score 3: [Score 3 Description]
Score 4: [Score 4 Description]
Score 5: [Score 5 Description]

</evaluation_criterion>

<output_format>
Provide your evaluation in the following structured format:

**Rationale:**
<Provide a comprehensive analysis of the survey's performance against the specific criterion. Include specific examples of strengths and weaknesses, with detailed justification for your assessment. Address how well the survey meets the criterion description and identify specific areas that align with or deviate from the scoring descriptions.>

**Final Score:**
<SCORE>X</SCORE>
(Where X is the score from 1 to 5 based on your evaluation)

Return your response in the following JSON format:
{
  "rationale": "Your detailed reasoning here",
  "score": X
}
</output_format>

Now conduct your comprehensive evaluation of the academic survey quality.
\end{tcblisting}

\begin{tcblisting}{
  listing only,
  listing engine=listings,
  enhanced, breakable,
  colback=blue!6, colframe=blue!60!black,
  arc=3mm, boxrule=0.6pt,
  left=6mm, right=6mm, top=4mm, bottom=4mm,
  title={Coverage Criterion}
}
Description: Coverage: Coverage assesses the extent to which the survey encapsulates all relevant aspects of the topic, ensuring comprehensive discussion on both central and peripheral topics.

Score 1: The survey has very limited coverage, only touching on a small portion of the topic and lacking discussion on key areas.  
Score 2: The survey covers some parts of the topic but has noticeable omissions, with significant areas either underrepresented or missing.  
Score 3: The survey is generally comprehensive in coverage but still misses a few key points that are not fully discussed.  
Score 4: The survey covers most key areas of the topic comprehensively, with only very minor topics left out.  
Score 5: The survey comprehensively covers all key and peripheral topics, providing detailed discussions and extensive information.  
\end{tcblisting}

\begin{tcblisting}{
  listing only,
  listing engine=listings,
  enhanced, breakable,
  colback=blue!6, colframe=blue!60!black,
  arc=3mm, boxrule=0.6pt,
  left=6mm, right=6mm, top=4mm, bottom=4mm,
  title={Structure Criterion}
}
Description: Structure: Structure evaluates the logical organization and coherence of sections and subsections, ensuring that they are logically connected.

Score 1: The survey lacks logic, with no clear connections between sections, making it difficult to understand the overall framework.  
Score 2: The survey has weak logical flow with some content arranged in a disordered or unreasonable manner.  
Score 3: The survey has a generally reasonable logical structure, with most content arranged orderly, though some links and transitions could be improved such as repeated subsections.  
Score 4: The survey has good logical consistency, with content well arranged and natural transitions, only slightly rigid in a few parts.  
Score 5: The survey is tightly structured and logically clear, with all sections and content arranged most reasonably, and transitions between adajecent sections smooth without redundancy.  
\end{tcblisting}

\begin{tcblisting}{
  listing only,
  listing engine=listings,
  enhanced, breakable,
  colback=blue!6, colframe=blue!60!black,
  arc=3mm, boxrule=0.6pt,
  left=6mm, right=6mm, top=4mm, bottom=4mm,
  title={Relevance Criterion}
}
Description: Relevance: Relevance measures how well the content of the survey aligns with the research topic and maintain a clear focus.

Score 1: The content is outdated or unrelated to the field it purports to review, offering no alignment with the topic.  
Score 2: The survey is somewhat on topic but with several digressions; the core subject is evident but not consistently adhered to.  
Score 3: The survey is generally on topic, despite a few unrelated details.  
Score 4: The survey is mostly on topic and focused; the narrative has a consistent relevance to the core subject with infrequent digressions.  
Score 5: The survey is exceptionally focused and entirely on topic; the article is tightly centered on the subject, with every piece of information contributing to a comprehensive understanding of the topic.  
\end{tcblisting}

\begin{tcblisting}{
  listing only,
  listing engine=listings,  % 原样显示
  enhanced, breakable,      % 可跨页
  colback=blue!6, colframe=blue!60!black,
  arc=3mm, boxrule=0.6pt,
  left=6mm, right=6mm, top=4mm, bottom=4mm,
  title={Survey-Arena Review Prompt}
}
# Paper 1:
Title: {title_1}
Figures: {figure_and_captions_1}
Content: {main_content_1}

# Paper 2:
Title: {title_2}
Figures: {figure_and_captions_2}
Content: {main_content_2}

You are provided with two survey papers on topic: {topic}.

As the area chair for a top ML conference, you can only select one paper. Start with a brief meta-review/reasoning of the pros and cons for each paper (two sentences), focusing on:

(1) insight and synthesis - moves beyond mere summarization to create new understanding and provides clear taxonomy;  
(2) thoroughness and accuracy - comprehensive coverage of literature with technical correctness;  
(3) structure and clarity - logical organization with compelling narrative;  
(4) scope and impact - well-defined scope with valuable future research directions;  
(5) presentation quality - professional polish, clear writing, and comprehensive evaluation of figures/tables presence and aesthetic quality.  

Be very critical and do not be biased by what the author claimed. Finally, provide your choice in a binary format.

**Your Task:**
1. Provide a detailed evaluation for Paper 1 using the above criteria.  
2. Provide a detailed evaluation for Paper 2 using the same criteria.  
3. Make a final decision by comparing the two papers and justifying your choice.  

STRICT OUTPUT INSTRUCTIONS:
- You MUST return a single valid JSON object.  
- Output ONLY JSON. No explanations, no Markdown, no code fences, no additional text before or after the JSON.  
- Use exactly these keys and types:  
  - "paper_1_review": string  
  - "paper_2_review": string  
  - "chosen_paper": "1" or "2"  
- Do NOT include any additional keys or trailing commas. If unsure, return empty strings for the review fields.  

Return JSON in exactly this shape:
{  
"paper_1_review": "Your meta-review and reasoning for paper 1",  
"paper_2_review": "Your meta-review and reasoning for paper 2",  
"chosen_paper": "1 or 2"  
}  

End your output immediately after the closing.
\end{tcblisting}

\newpage
\subsection{Comparison between AutoSurvey and IterSurvey. }
\label{subsec:Comparison_AutoSurvey_IterSurvey}
\begin{figure}[h]
  \centering
  \begin{subfigure}[t]{0.47\linewidth} 
    \centering
    \includegraphics[width=0.9\linewidth]{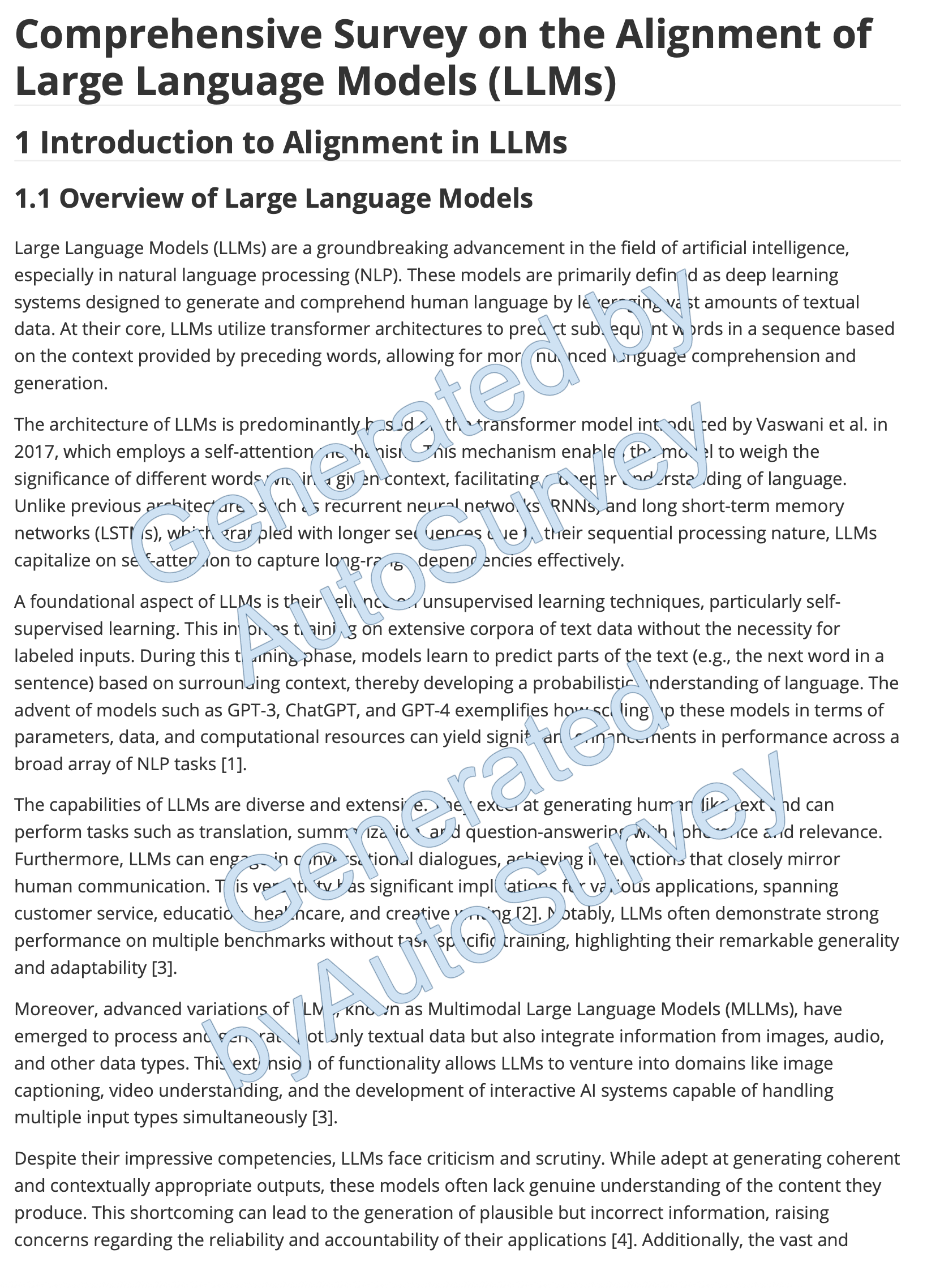} 
    \par\medskip
    \includegraphics[width=0.9\linewidth]{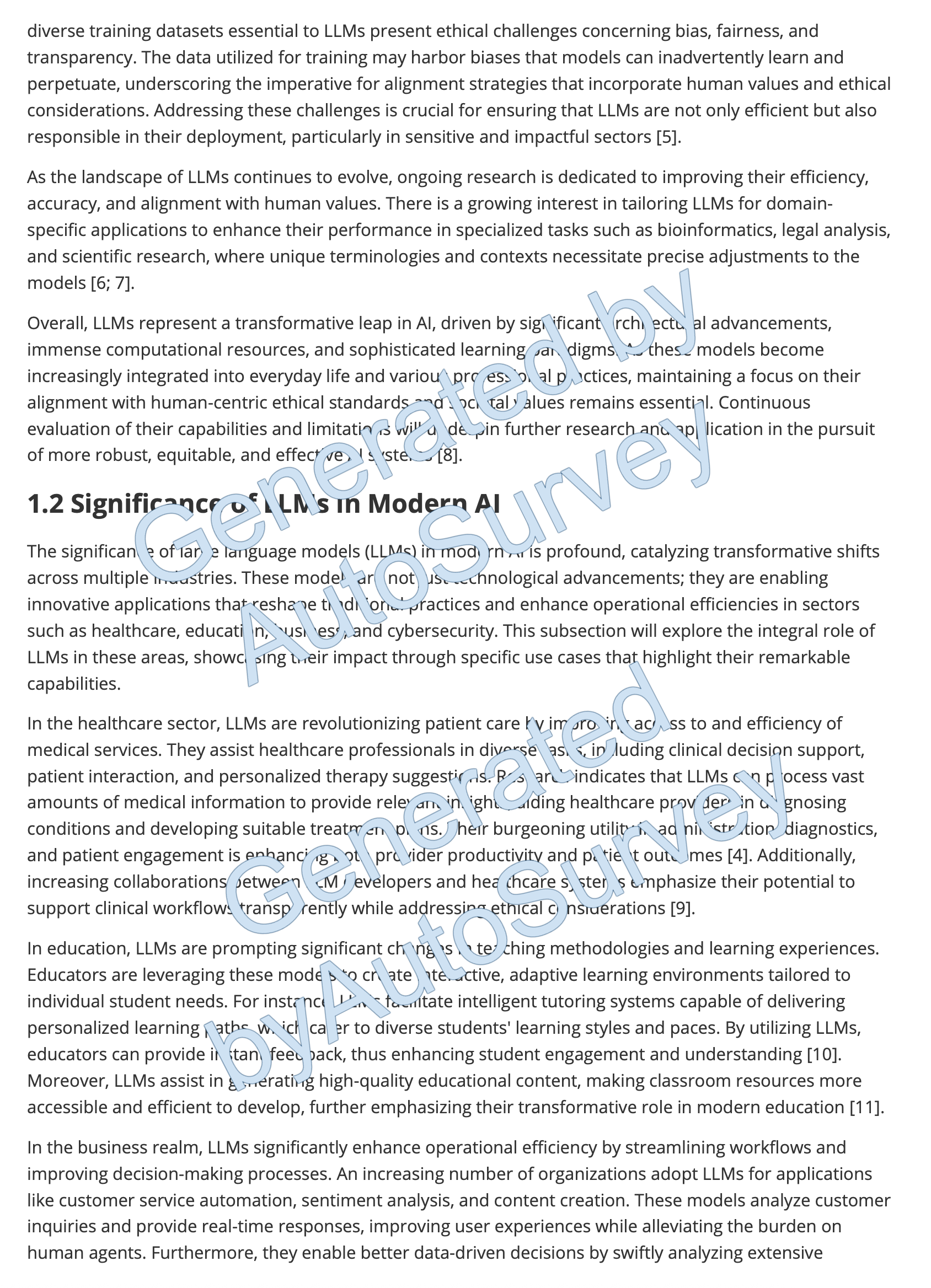} 
    \label{fig:autosurvey-v1-1} 
  \end{subfigure}
  \hfill
  \begin{subfigure}[t]{0.47\linewidth} 
    \centering
    \includegraphics[width=0.9\linewidth]{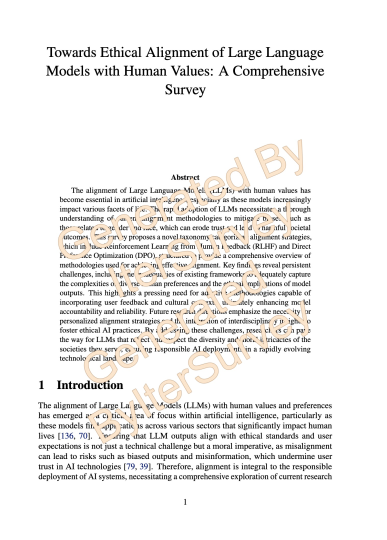} 
    \par\medskip
    \includegraphics[width=0.9\linewidth]{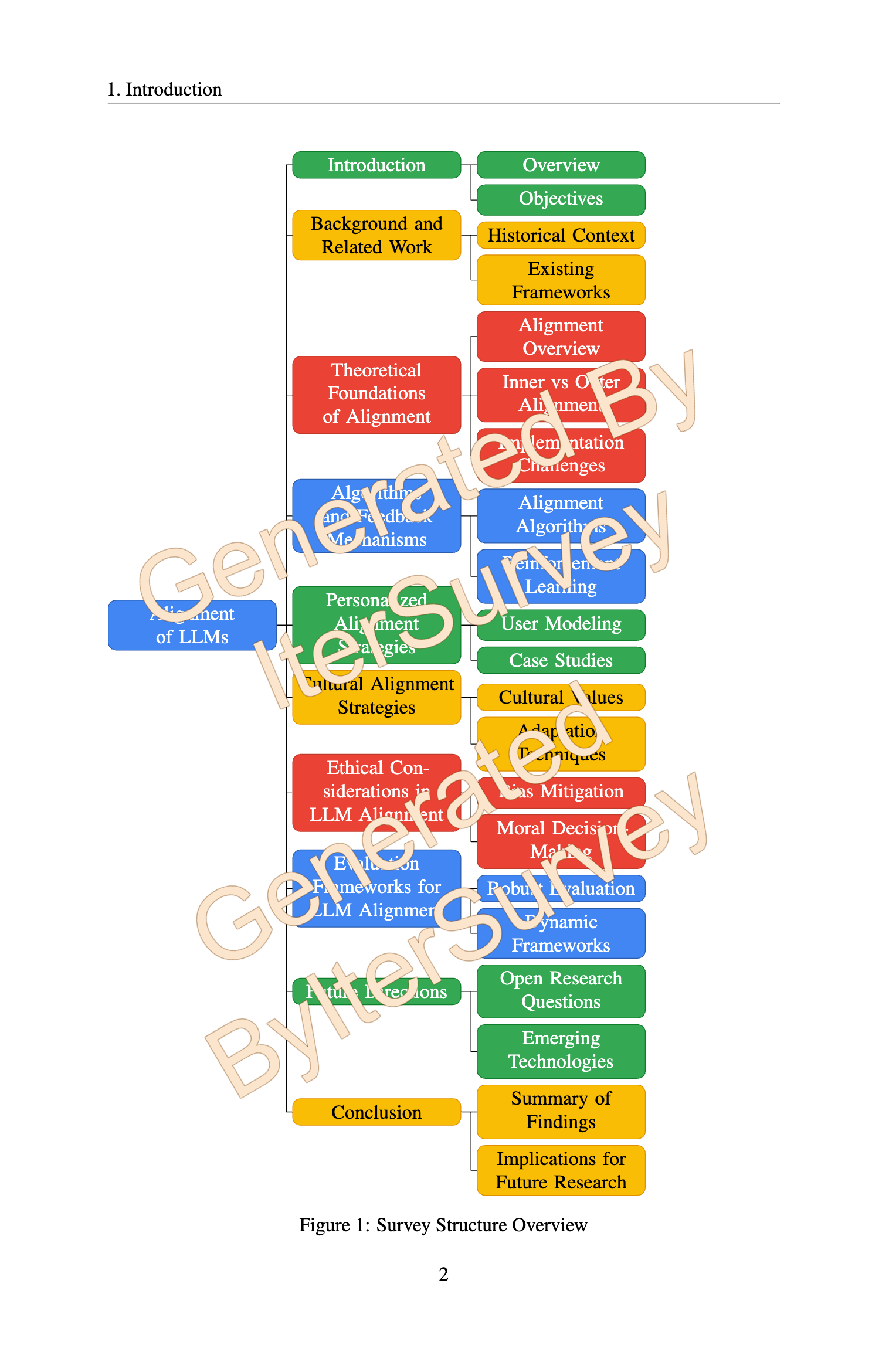} 
    \label{fig:autosurvey-v2-1} 
  \end{subfigure}

  \caption{LLM-generated survey comparison between AutoSurvey and IterSurvey. }
  \label{fig:autosurvey-compare-2x2-1}
\end{figure}

\begin{figure}[t]
  \centering
  \begin{subfigure}[t]{0.49\linewidth}
    \centering
    \includegraphics[width=0.96\linewidth]{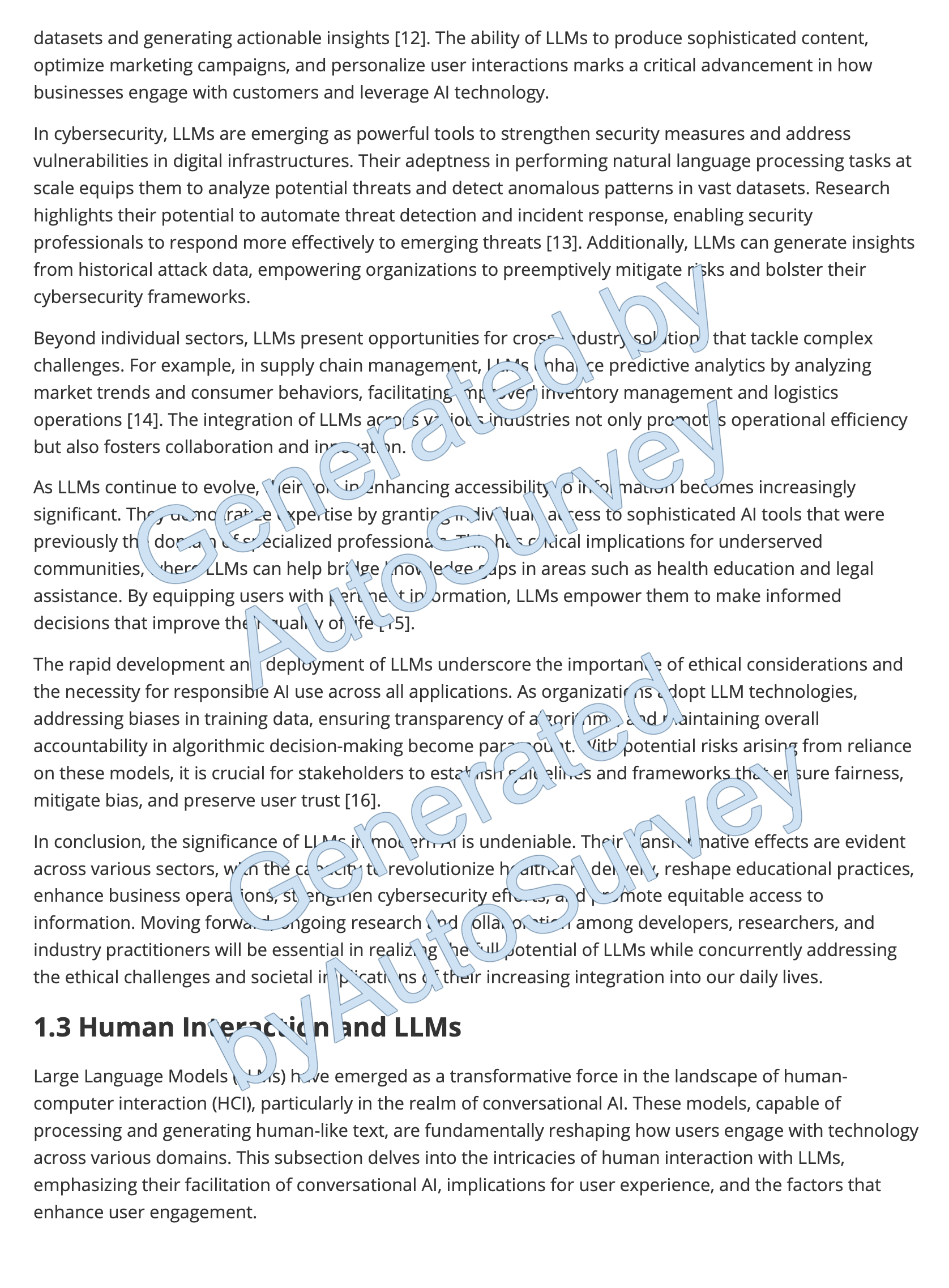}
    \par\medskip
    \includegraphics[width=0.96\linewidth]{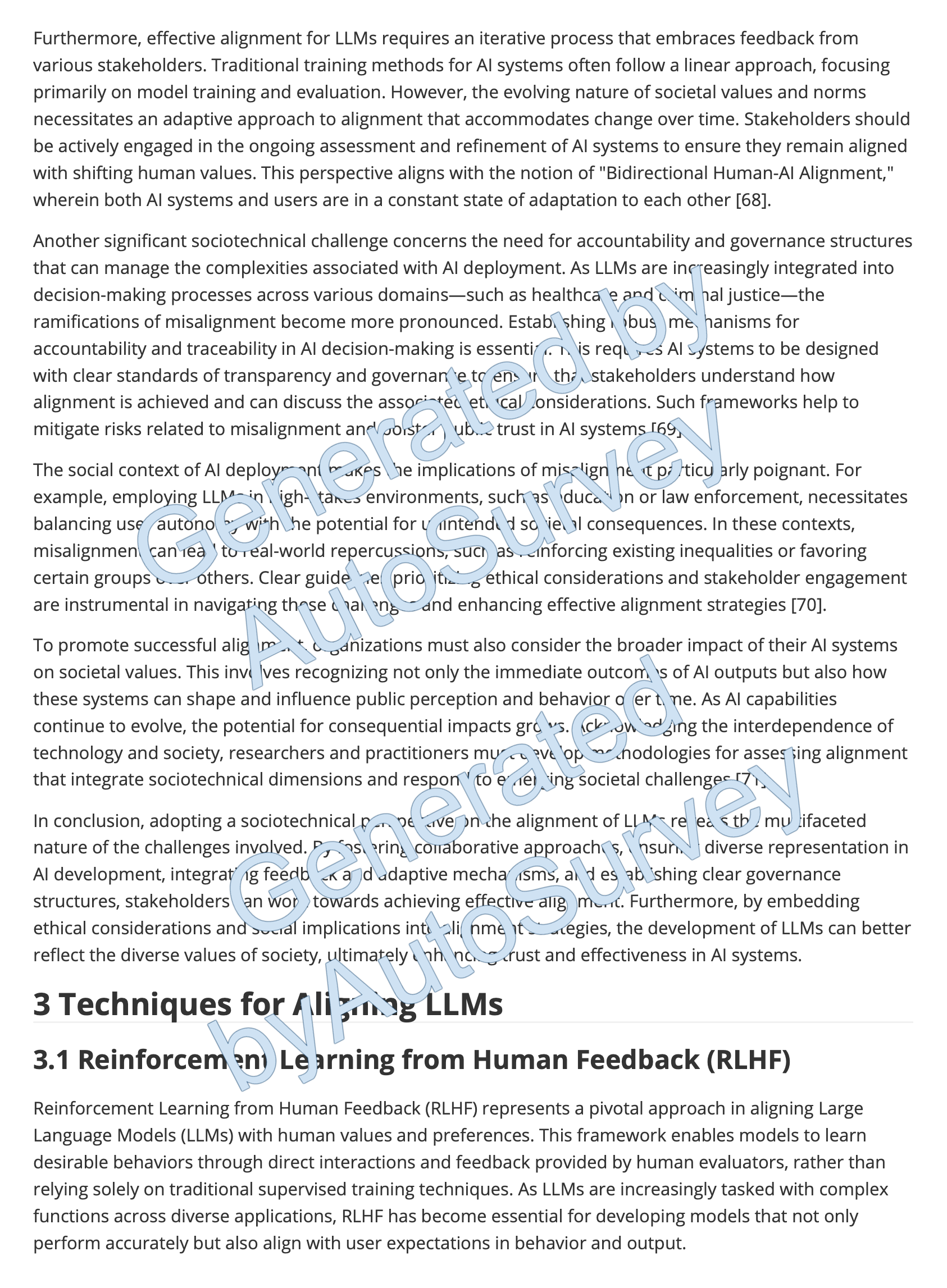}
    
    \label{fig:autosurvey-v1}
  \end{subfigure}
  \hfill
  \begin{subfigure}[t]{0.49\linewidth}
    \centering
    \includegraphics[width=0.96\linewidth]{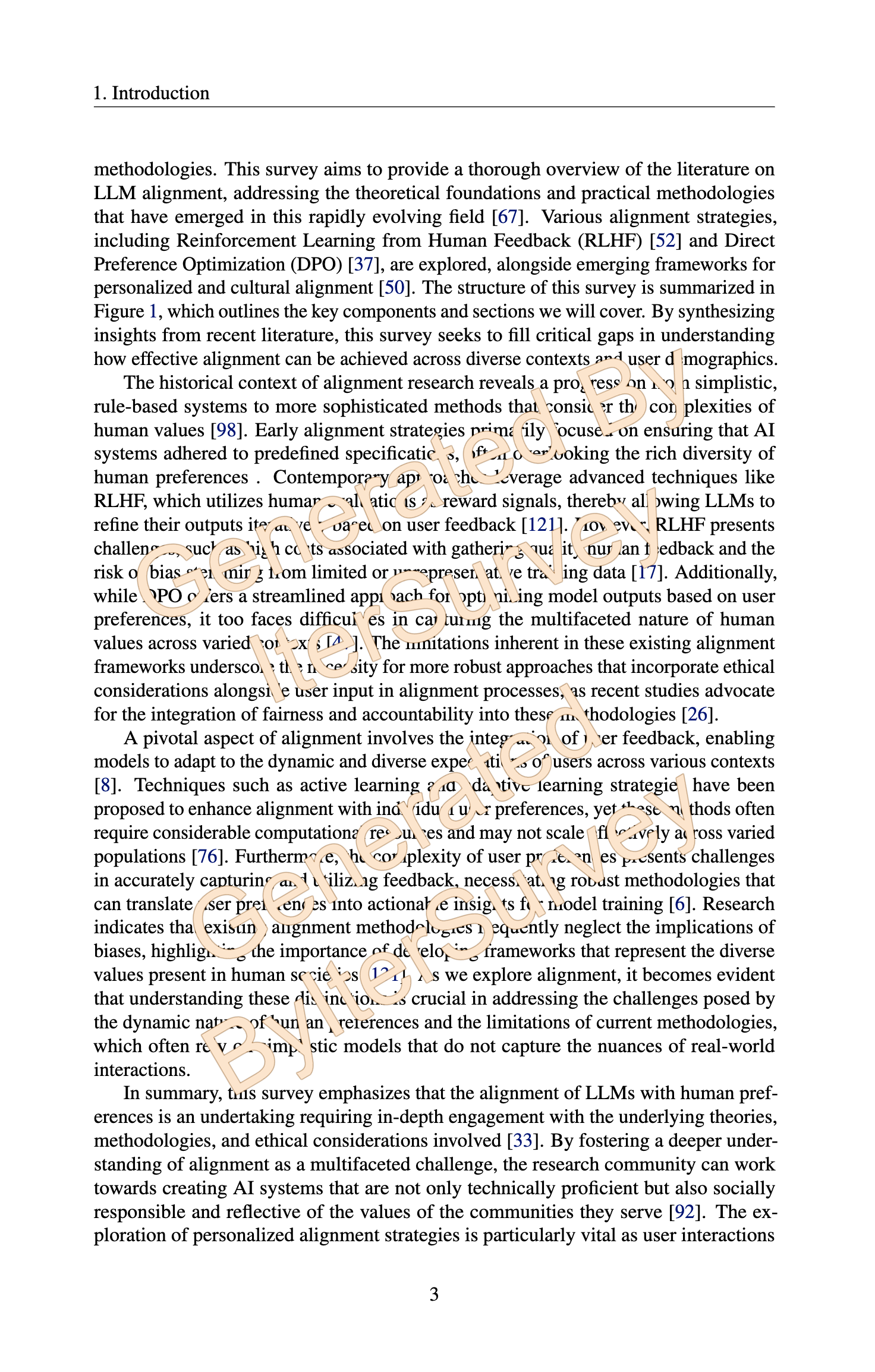}
    \par\medskip
    \includegraphics[width=0.96\linewidth]{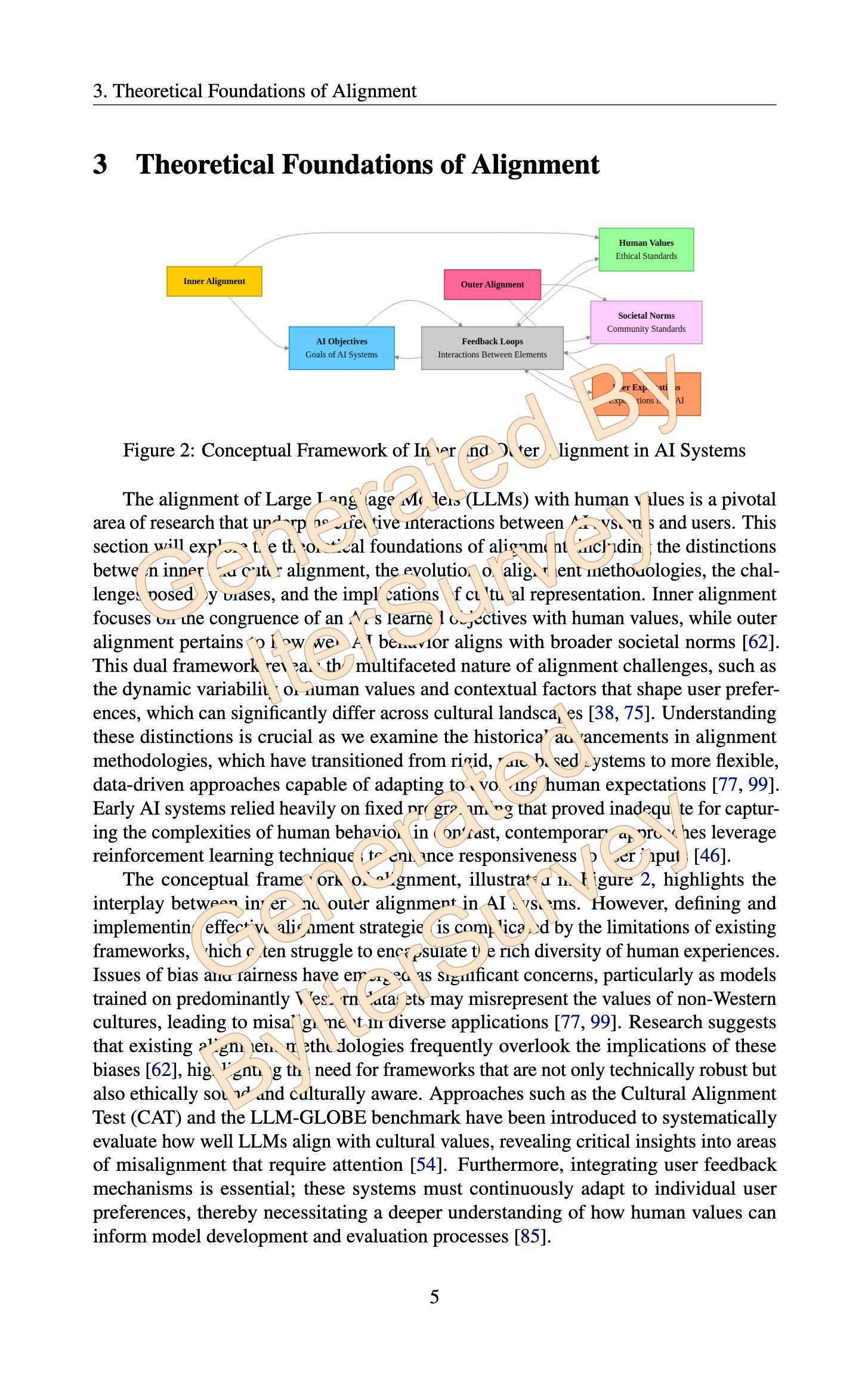}
    \label{fig:autosurvey-v2}
  \end{subfigure}

  \caption{LLM-generated survey comparison between AutoSurvey and IterSurvey. }
  \label{fig:autosurvey-compare-2x2-2}
\end{figure}

\begin{figure}[t]
  \centering
  \setlength{\tabcolsep}{4pt} 
  \begin{tabular}{@{} m{0.5\linewidth} m{0.5\linewidth} @{}}
    \subcaptionbox{AutoSurvey\label{fig:autosurvey-v1}}[\linewidth]{%
      \begin{minipage}[t]{\linewidth}
        \centering
        \includegraphics[width=0.9\linewidth]{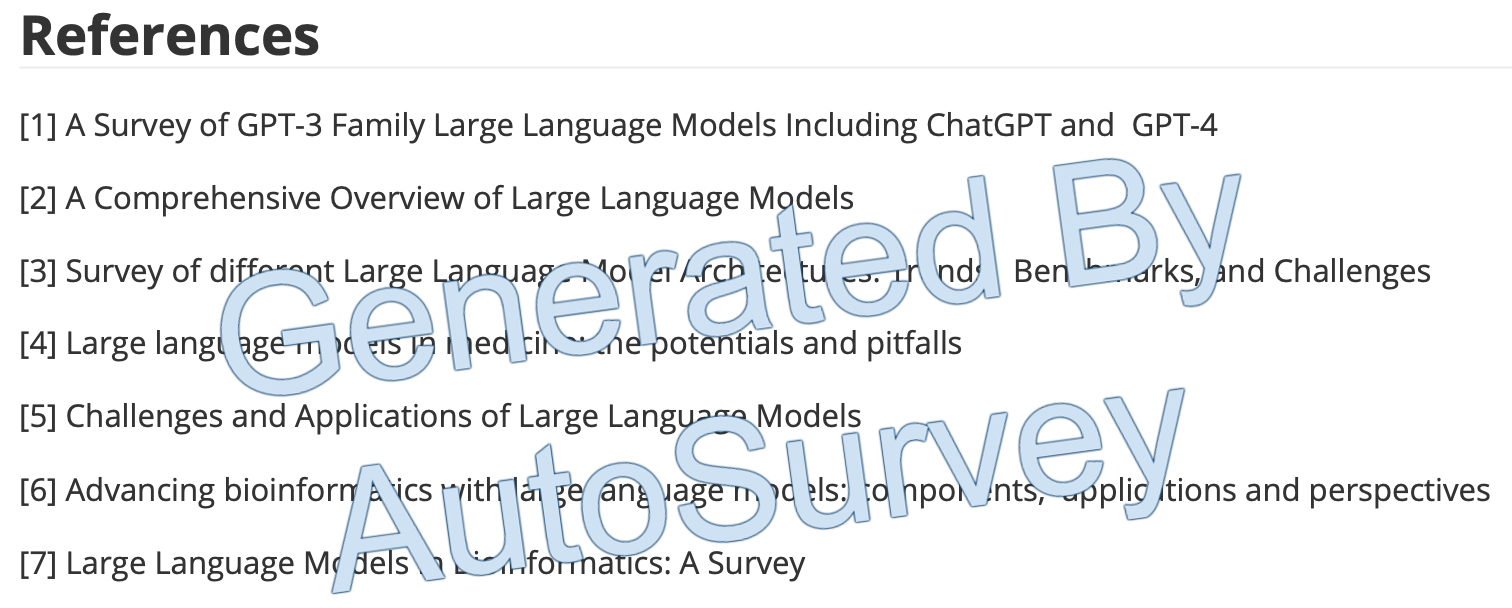}
        \par\vspace{3pt}
        \includegraphics[width=0.9\linewidth]{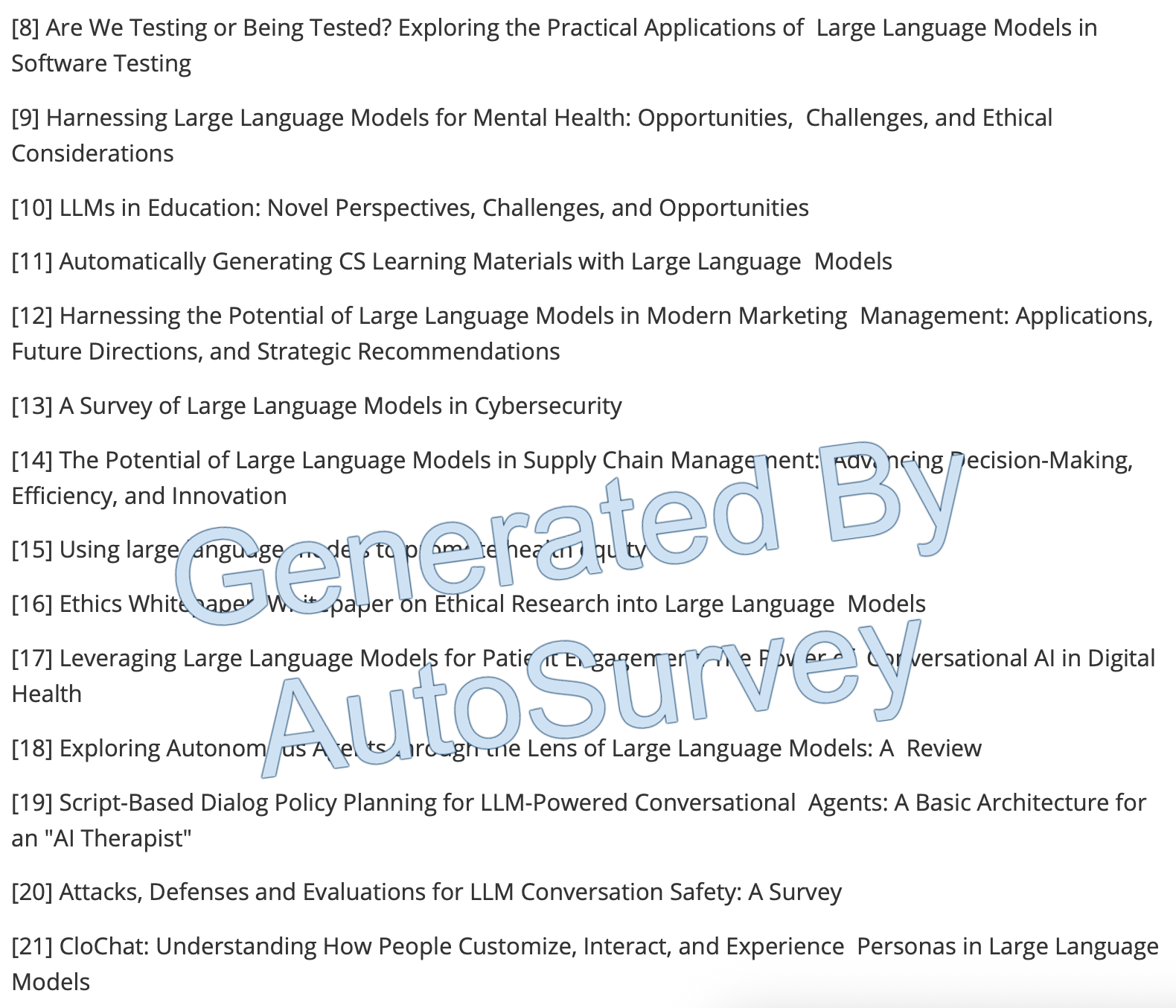}
      \end{minipage}
    }
    &
    \subcaptionbox{IterSurvey\label{fig:autosurvey-v2}}[\linewidth]{%
      \centering
      \includegraphics[width=\linewidth]{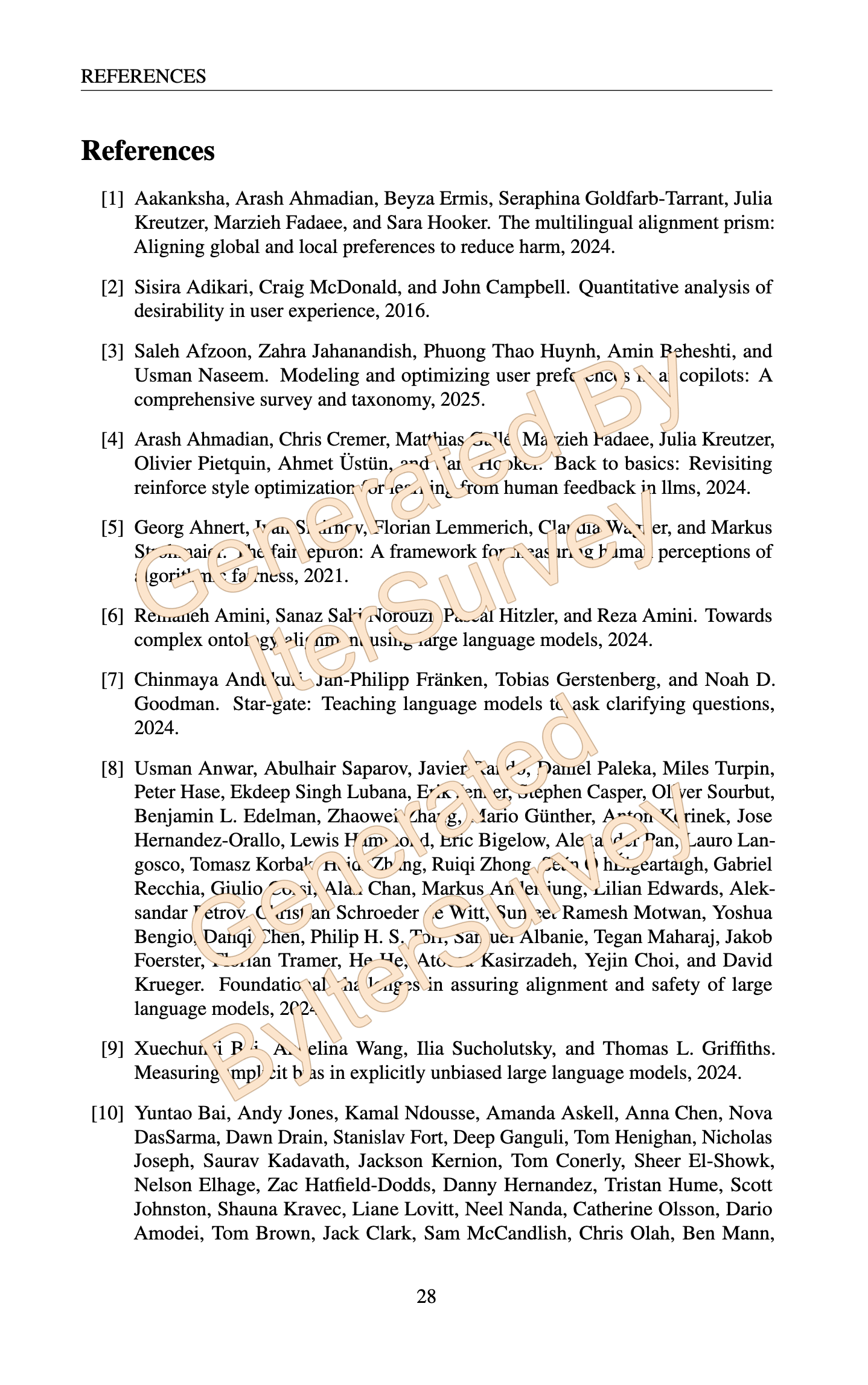}
    }
  \end{tabular}
  \caption{LLM-generated survey comparison between AutoSurvey and IterSurvey. }
  \label{fig:autosurvey-compare-left2-right1}
\end{figure}

\begin{figure}[t]
  \centering
  \begin{subfigure}[t]{0.49\linewidth}
    \centering
    \includegraphics[width=\linewidth]{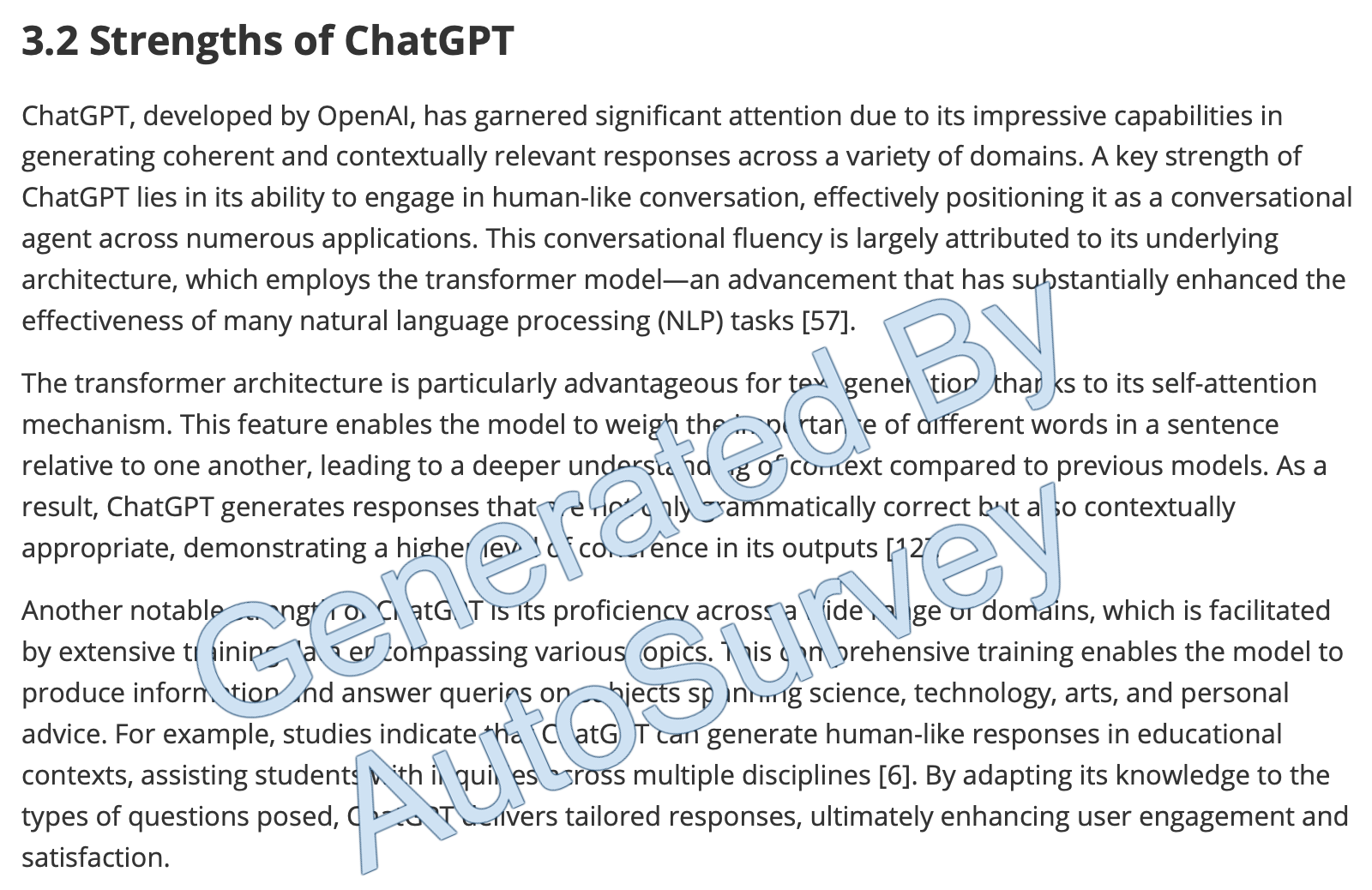}
    \caption{AutoSurvey}
    \label{fig:autosurvey-v1}
  \end{subfigure}
  \hfill
  \begin{subfigure}[t]{0.49\linewidth}
    \centering
    \includegraphics[width=\linewidth]{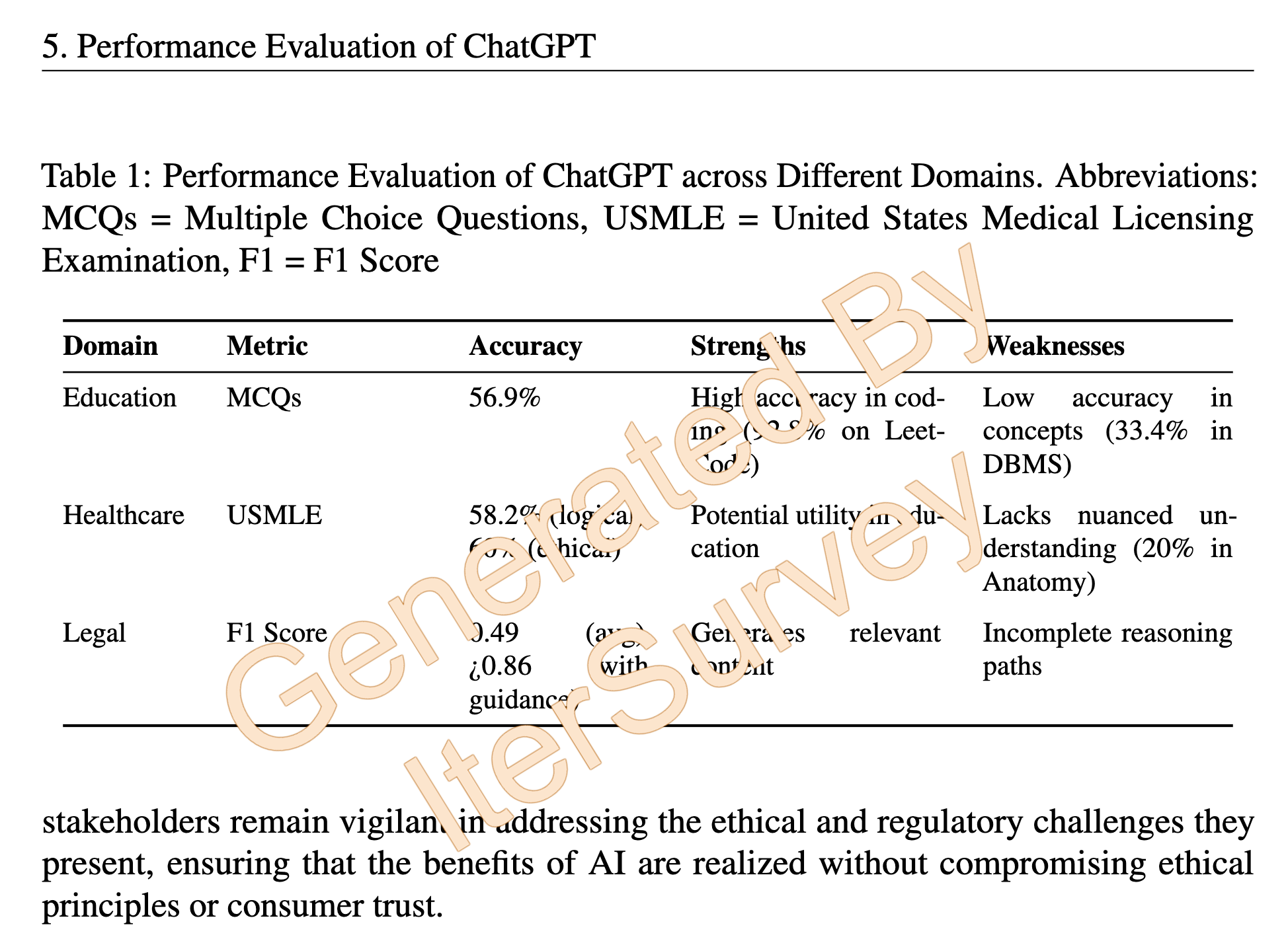}
    \caption{IterSurvey}
    \label{fig:autosurvey-v2}
  \end{subfigure}
  \caption{LLM-generated survey comparison between AutoSurvey and IterSurvey. }
  \label{fig:autosurvey-compare}
\end{figure}

\end{document}